\documentclass[journal]{IEEEtran}
\ifCLASSINFOpdf
\else
\fi
\usepackage{cite}
\usepackage{amsmath,amssymb,amsfonts}
\usepackage{algorithmic}
\usepackage{graphicx}
\usepackage{textcomp}
\usepackage{subfig}
\usepackage{soul}
\soulregister\cite7 % Õë¶Ô\citeÃüÁî
\soulregister\citep7 % Õë¶Ô\citepÃüÁî
\soulregister\citet7 % Õë¶Ô\citetÃüÁî
\soulregister\ref7 % Õë¶Ô\refÃüÁî
\soulregister\pageref7 % Õë¶Ô\pagerefÃüÁî
\begin{document}
\title{Medical Image Segmentation with Limited Supervision: A Review of Deep Network Models}
\author{Jialin~Peng, and Ye~Wang% <-this % stops a space
\thanks{J. Peng was with College of Computer Science and Technology, Huaqiao University, Xiamen 361021, China}
\thanks{Y. Wang was with School of Economics and Finance, Huaqiao University, Xiamen 361021, China}
\thanks{Corresponding author: Jialin Peng (e-mail: 2004pjl@163.com).}}
\maketitle
\begin{abstract}
Despite  the remarkable performance of deep learning methods on various tasks, most cutting-edge models rely  heavily on large-scale annotated training examples, which are often unavailable for clinical and health care tasks. The labeling costs for medical images are very high, especially in medical image segmentation, which typically requires intensive pixel/voxel-wise labeling. Therefore, the strong capability of learning and generalizing from limited supervision, including a limited amount of annotations, sparse annotations, and inaccurate annotations, is crucial for the successful application of deep learning models in medical image segmentation. However, due to its intrinsic difficulty, segmentation with limited supervision is challenging and specific model  design and/or learning strategies are needed. In this paper, we provide a systematic and up-to-date review of the solutions above, with summaries and comments about the methodologies.  We also highlight several problems in this field, discussed future directions  observing further investigations.
\end{abstract}

\begin{IEEEkeywords}
Medical image segmentation, semi-supervised segmentation, partially-supervised segmentation, noisy label, sparse annotation
\end{IEEEkeywords}

%\titlepgskip=-15pt
\IEEEpeerreviewmaketitle

\section{Introduction}
\label{sec:1}
\IEEEPARstart{M}{edical} image segmentation, identifying the pixels/voxels of  anatomical
or pathological structures from background biomedical images, is of vital importance in many biomedical applications, such as computer-assisted  diagnosis,  radiotherapy planning, surgery simulation, treatment, and follow-up of many diseases.  Typical medical image segmentation tasks include brain and tumor segmentation \cite{dora2017state,luo2020hdc,wang2017automatic},  cardiac segmentation \cite{chen2020deep}, liver and tumor segmentation \cite{li2018h,peng2014liver,peng20153d,hu2016automatic},  cell and subcellular structures \cite{deng2020deep,peng2019mitochondria,peng2020unsupervised}, multi-organ segmentation \cite{cerrolaza2019computational} and  lung  and pulmonary nodules \cite{shi2020review}, vessel segmentation \cite{moccia2018blood}, etc., and thus can deliver crucial information about the objects of interest.  While semantic segmentation of medical images  involves labeling each pixel/voxel with the semantic class,  instance segmentation (such as cell segmentation) extends semantic segmentation to discriminate each instance within the same class. Recently, deep learning methods have achieved impressive performance improvements on various medical image segmentation tasks and set the new state of the art.  Numerous image segmentation algorithms have
been developed in the literature and have made great progress on the designs and performance of deep network models \cite{litjens2017survey,taghanaki2020deep}.

  \begin{figure*}[!t]
    \centering
    \includegraphics[width=0.85\textwidth]{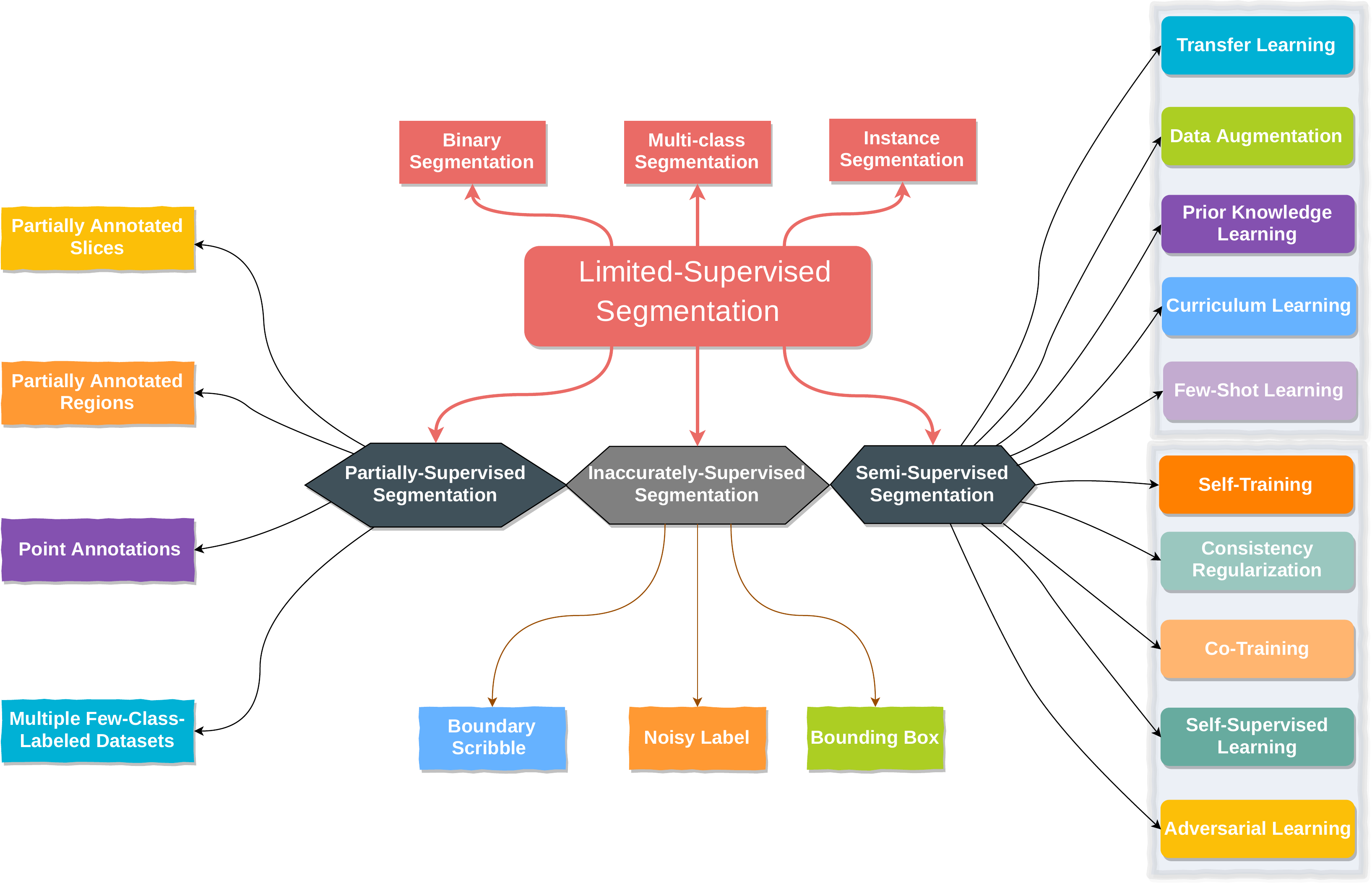}
    \caption{A taxonomy of medical image segmentation under limited supervision.     }
  \label{fig:1}
  \end{figure*}
However, the scarcity of high-quality  annotated training data has been a significant challenge for medical image segmentation. The strong generalization capabilities of most  cutting-edge segmentation models, which are usually deep and wide networks, highly rely on large-scale and high-quality pixel-wise annotated data, which are often unavailable for clinical and health care tasks. In fact, it is an expensive and time-consuming process to manually annotate medical images at pixel-level since it requires the knowledge of experienced clinical experts. The scarcity of annotated medical imaging data is further exacerbated by the data differences  in patient populations, acquisition parameters and protocols, sequences, vendors, and centers, which may result in obvious statistical shifts. Thus, it is even challenging to collect a sufficiently large number of training data due to the  heterogeneous nature of medical imaging data and the strict legal and ethical requirements for patient privacy. The data scarcity problem is much more severe for emerging tasks and new environments, where quick model employment is expected. However, only a limited amount of annotations with limited quality are available. Therefore, the high  cost of pixel-level labeling and the privacy and security of data hinder the model training and their scalability to novel images of emerging tasks and new environments, which subsequently  hamper the application of deep segmentation models in real-world clinical and health care usage.  Thus, learning strong and robust segmentation models from limited labeled data and  readily available  unlabeled data  is crucial for the successful application of deep learning models in clinical usage and health care.

% Another significant challenge is the data mismatch, concerning the domain shift between the data used for model training and  real-world clinical data due to different patient populations, sample selection bias, different scanning protocols and parameters, and combinations thereof.
 These challenges have  inspired
many research efforts on learning with limited supervision, where the training data only have a limited amount of annotated examples, accurate but sparse annotations, inaccurate annotations, coarse-level annotations,  and combinations of thereof. However, due to its intrinsic difficulty, segmentation with limited supervision is challenging,  and specific model  design and/or learning strategies are needed. Despite these challenges, researchers  have introduced a diverse set of deep network models \cite{taghanaki2020deep} that can handle incomplete, sparse, inaccurate, or coarse annotations. However, the progress is more slowly  than that of fully supervised learning. In this paper, we will take a systematic and up-to-date  look at the development of recent technologies that explored the unlabeled examples and prior knowledge to address the limited supervision and small data problem.

Several comprehensive surveys exist  about the deep learning methods \cite{litjens2017survey,zhou2020review} or its subcategories such as generative adversarial networks (GAN) \cite{yi2019generative} for general medical image analysis (including classification, reconstruction, detection, registration, and segmentation) \cite{litjens2017survey,zhou2020review,altaf2019going} or a specialized topic \cite{dora2017state,chen2020deep,deng2020deep,cerrolaza2019computational,shi2020review}, such as digital pathology image analysis \cite{cerrolaza2019computational,srinidhi2020deep}, and incorporating domain knowledge \cite{xie2020survey}. While several reviews have reviewed the application of deep learning for a specific application of the segmentation, such as cardiac image segmentation \cite{chen2020deep}, brain tissue segmentation \cite{dora2017state}, brain tumor segmentation \cite{wadhwa2019review}, and segmentation for covid-19 \cite{shi2020review}, the surveys by \cite{heimann2009comparison,taghanaki2020deep,minaee2020image} review advancement of deep network architectures, losses,  and training strategies for medical image segmentation. There are also several reviews closely related to our paper. Karimi \textit{et al.} \cite{karimi2020deep} reviewed deep learning methods dealing with label noises for medical image analysis, where most of the representative studies are about medical image classification. Zhang \textit{et al.} \cite{zhang2019survey} provided a review of deep learning methods that tackling small sample problems for various medical image analysis tasks, such as classification, detection, and segmentation. The most relevant survey to our study is  \cite{tajbakhsh2020embracing}, which focused on deep learning solutions for medical image segmentation with an imperfect training set. The current survey focuses  on deep network models for medical image segmentation with limited supervision and provides a more updated review of recent advancements. An overview of  the main body of
this survey is demonstrated in Fig. \ref{fig:1}.

%While existing works have explored a variety of techniques to push the envelop of weakly-supervised semantic segmentation, there is still a significant gap compared to the supervised methods.

To sum up, the main contributions of this paper are:
\begin{itemize}
  \item We provide a systematic and up-to-date review of medical image segmentation with limited supervision. One can quickly identify the frontier ideas in this field and, more importantly obtain an overall  picture of the problems and methodologies in this research area.
  \item We categorize the problem of limited-supervised segmentation  into semi-supervised segmentation, partially supervised segmentation, and inaccurately-supervised segmentation and offer a structural review of recent advances in methods that can be used to address these problems. We also offer summaries and comments  about the pros and cons of the methodologies in each category, and the connections of methods in  different categories.
  \item We also highlight  several problems in this field and discuss the limitation and  the new trends and future directions for medical image segmentation with limited supervision.
\end{itemize}

The paper is organized as follows. In Section \ref{sec:2}, we provide  preliminary knowledge about medical image segmentation and the basic deep network architectures for this task, as well as the categorization of medical image segmentation with limited supervision.  Section \ref{sec:3} to Section \ref{sec:5} provide a detailed review of methods for semi-supervised segmentation, partially-supervised segmentation, and inaccurately supervised segmentation, respectively.  Section \ref{sec:6} discusses future
directions.  Lastly, Section \ref{sec:6} concludes this survey.

 \begin{figure}[!t]
    \centering
   % \subfloat[EM image]{\includegraphics[width=0.155\textwidth]{test015.pdf}} \hspace{0.01mm}
%    \subfloat[Semantic seg.]{\includegraphics[width=0.155\textwidth]{GT3.pdf}} \hspace{0.01mm}
%    \subfloat[Instance seg.]{\includegraphics[width=0.155\textwidth]{instance3.pdf}}\\
      \subfloat[Original image]{\includegraphics[width=0.157\textwidth]{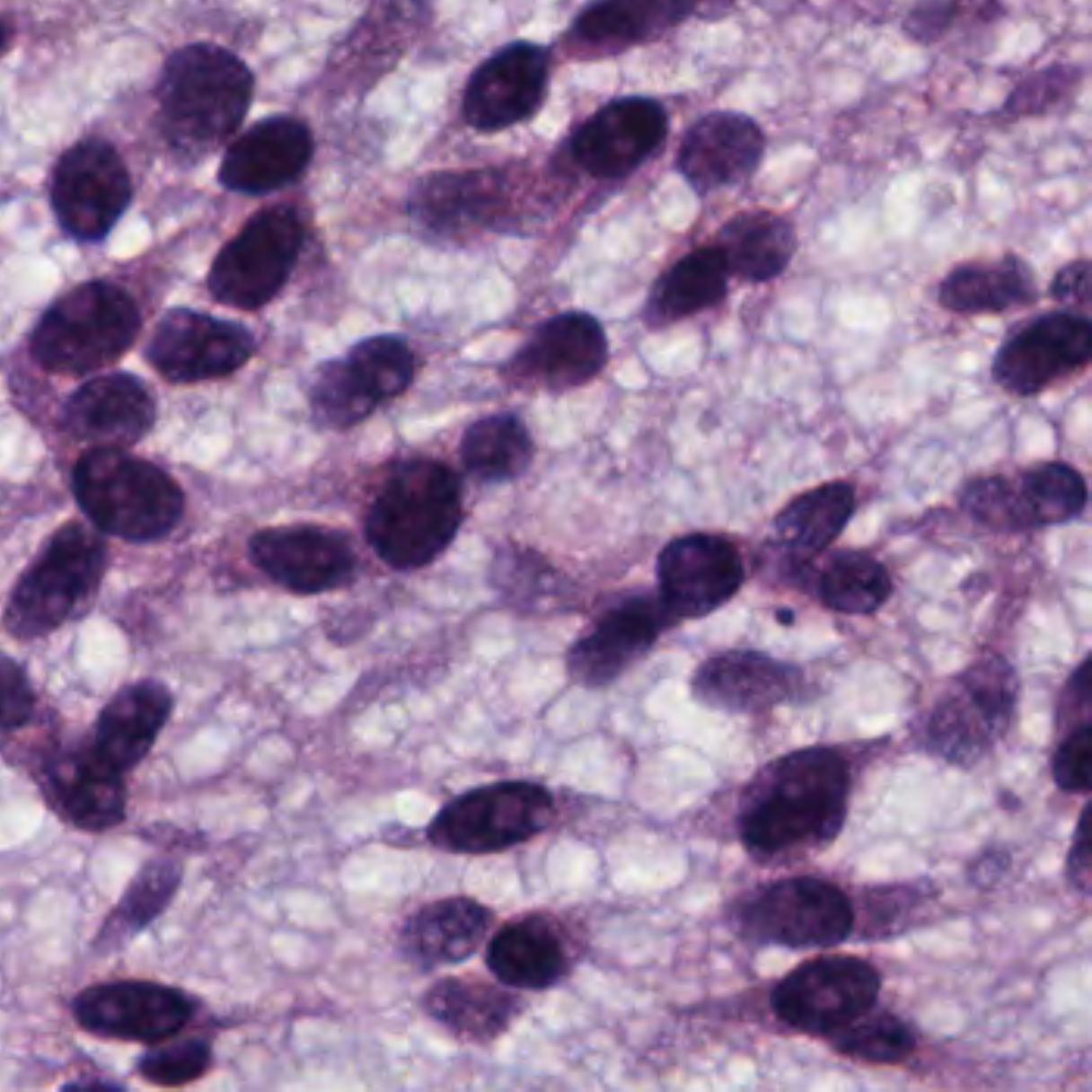}}\hspace{0.001mm}
    \subfloat[Semantic seg.]{\includegraphics[width=0.157\textwidth]{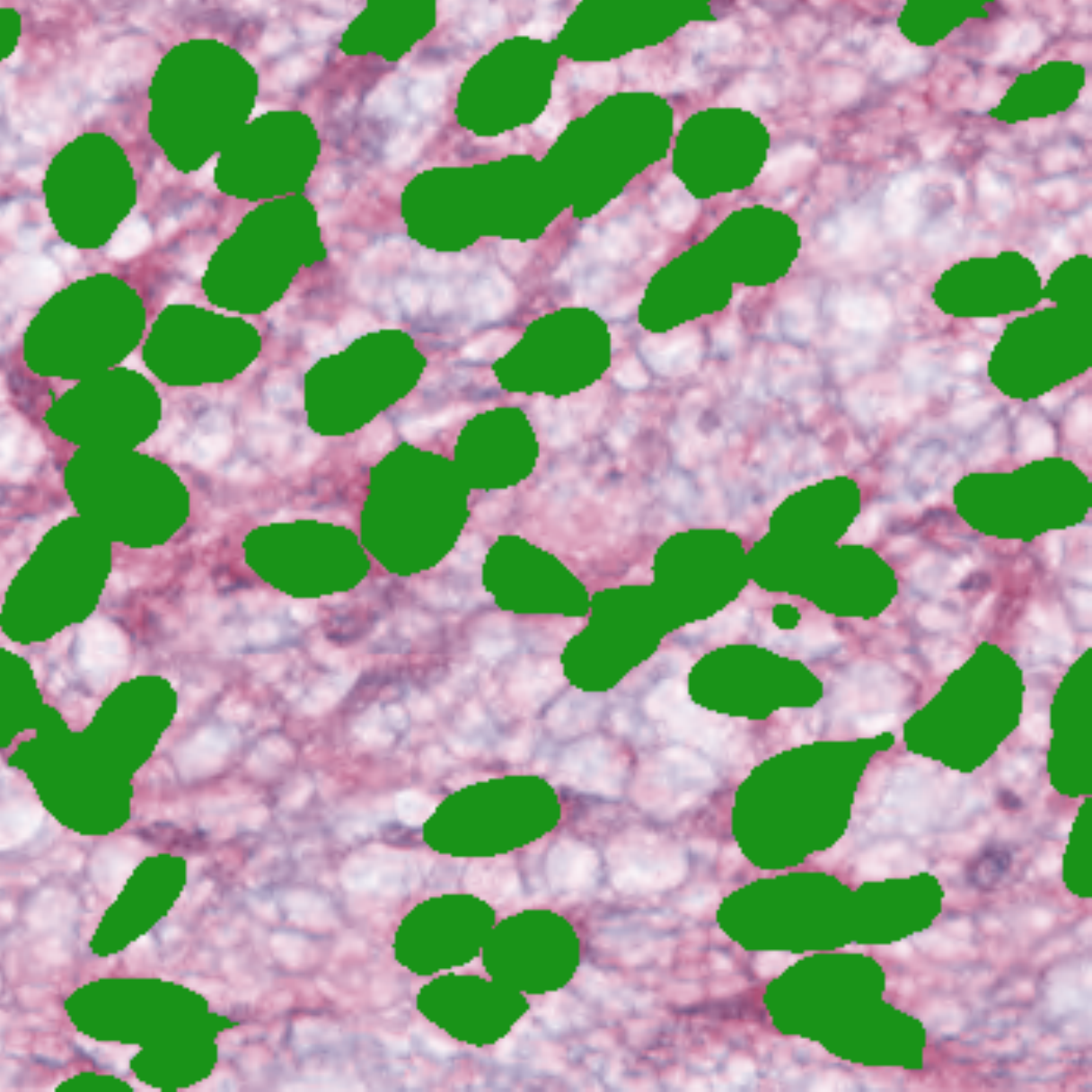}} \hspace{0.001mm}
    \subfloat[Instance seg.]{\includegraphics[width=0.157\textwidth]{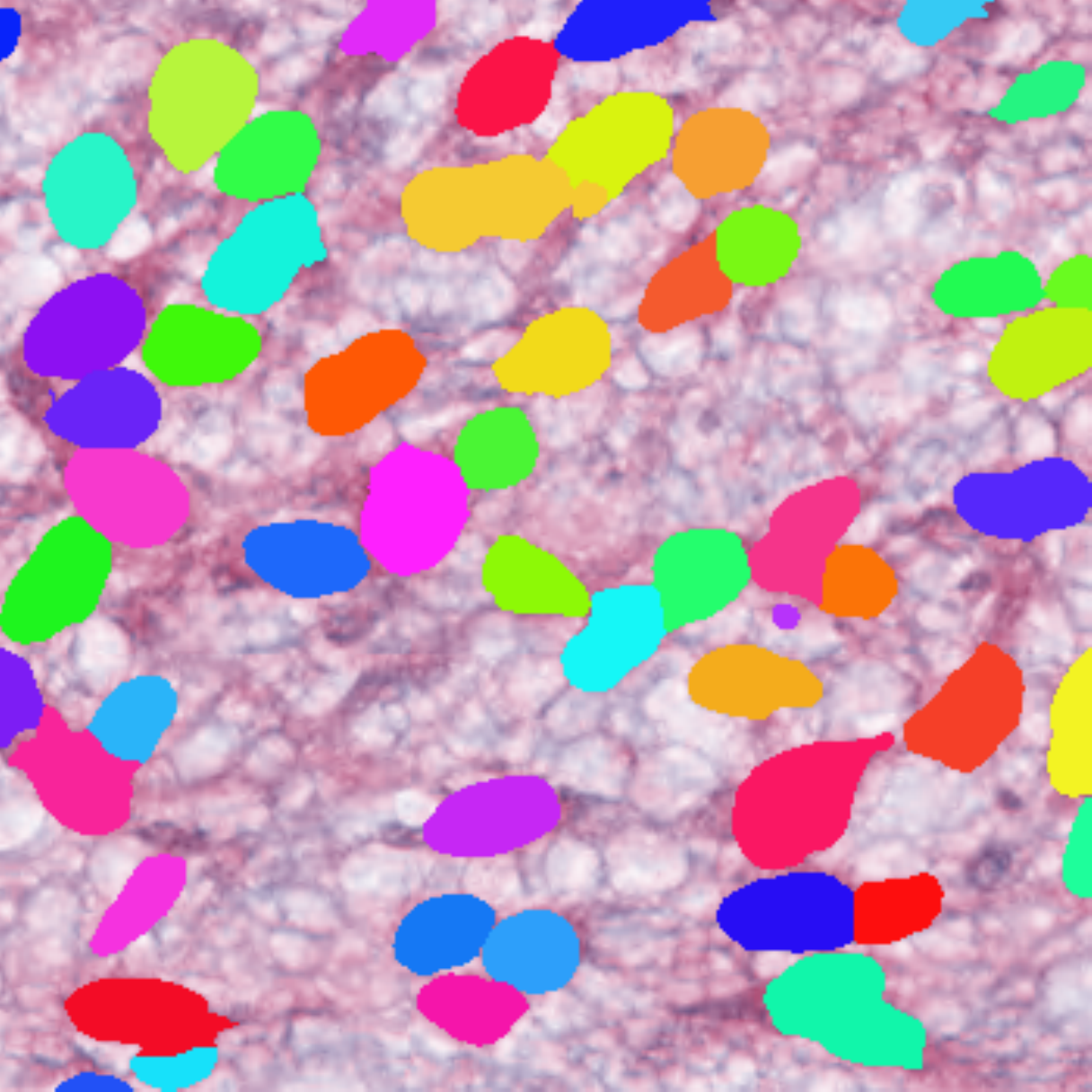}}
 %   \subfloat[a]{\includegraphics[width=0.48\textwidth]{StudentTeacher10.png}}
    \caption{A comparison of b) semantic segmentation and c) instance segmentation of nuclei from a) histopathology  images. While semantic segmentation labels all nuclei pixels as the foreground, instance segmentation associates each nucleus with a unique label. Different labels are denoted by different colors. }
  \label{fig:0}
  \end{figure}

\section{Overview} \label{sec:2}
\textbf{Medical image segmentation} involves delineating the anatomical
or pathological structures from  medical images of various modalities. As pointed out in \cite{zhou2020review}, medical images are heterogonous with imbalanced classes and  have multiple modalities with sparse annotations. Thus, it is complicated and challenging to analyze various  medical images. Here, we focus on the medical image segmentation problem, which typically consists of semantic segmentation and instance segmentation. Semantic segmentation refers to the task of assigning each pixel/voxel with a semantic category label (such as liver, kidney, etc.). Thus, semantic segmentation generates per-pixel segmentation masks, and the multiple objects  of the same category are treated  as one entity. In contrast, instance segmentation delineates the instances of each category. An illustration of the difference of  semantic segmentation  and instance segmentation is  exemplified in Fig. \ref{fig:0}. For semantic segmentation of nuclei, all nuclei pixels are annotated with the same label, and instance segmentation associates different nuclei with different labels.

\begin{figure}[!t]
    \centering
  \includegraphics[width=0.42\textwidth]{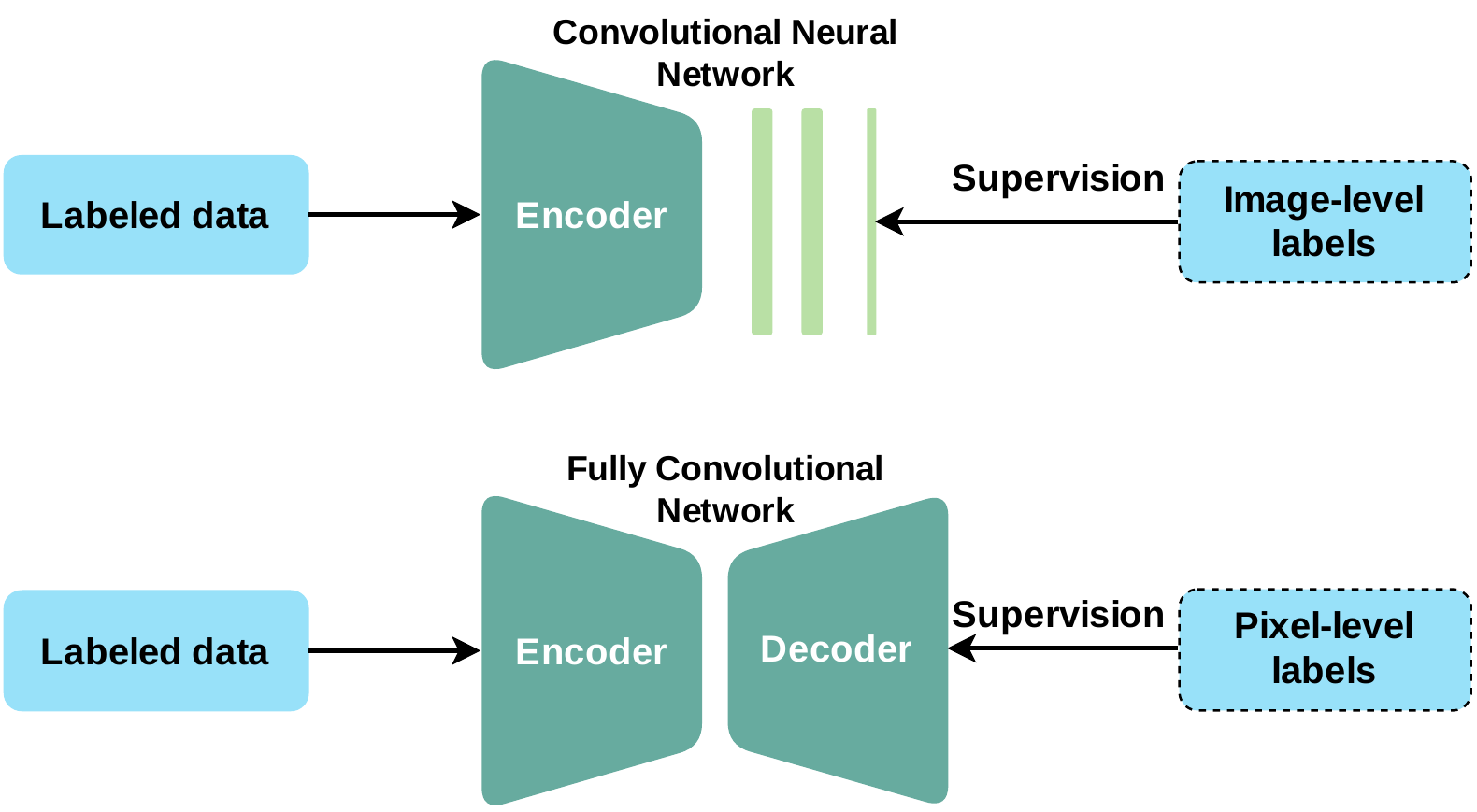}
   % \subfloat[GAN ]{\includegraphics[width=0.44\textwidth]{GAN.pdf}}
 %   \subfloat[a]{\includegraphics[width=0.48\textwidth]{StudentTeacher10.png}}
    \caption{A comparison of standard convolutional neural networks (CNNs) (Top) and fully convolutional networks (FCNs) (Bottom) in encoder-decoder architecture. An FCN is a CNN that substitutes the the  fully connected layers in the standard CNN  by  convolution layers and deconvolution/transposed convolution layers.}
  \label{fig:01}
  \end{figure}

\begin{figure}[!t]
    \centering
      \subfloat[U-Net]{\includegraphics[width=0.5\textwidth]{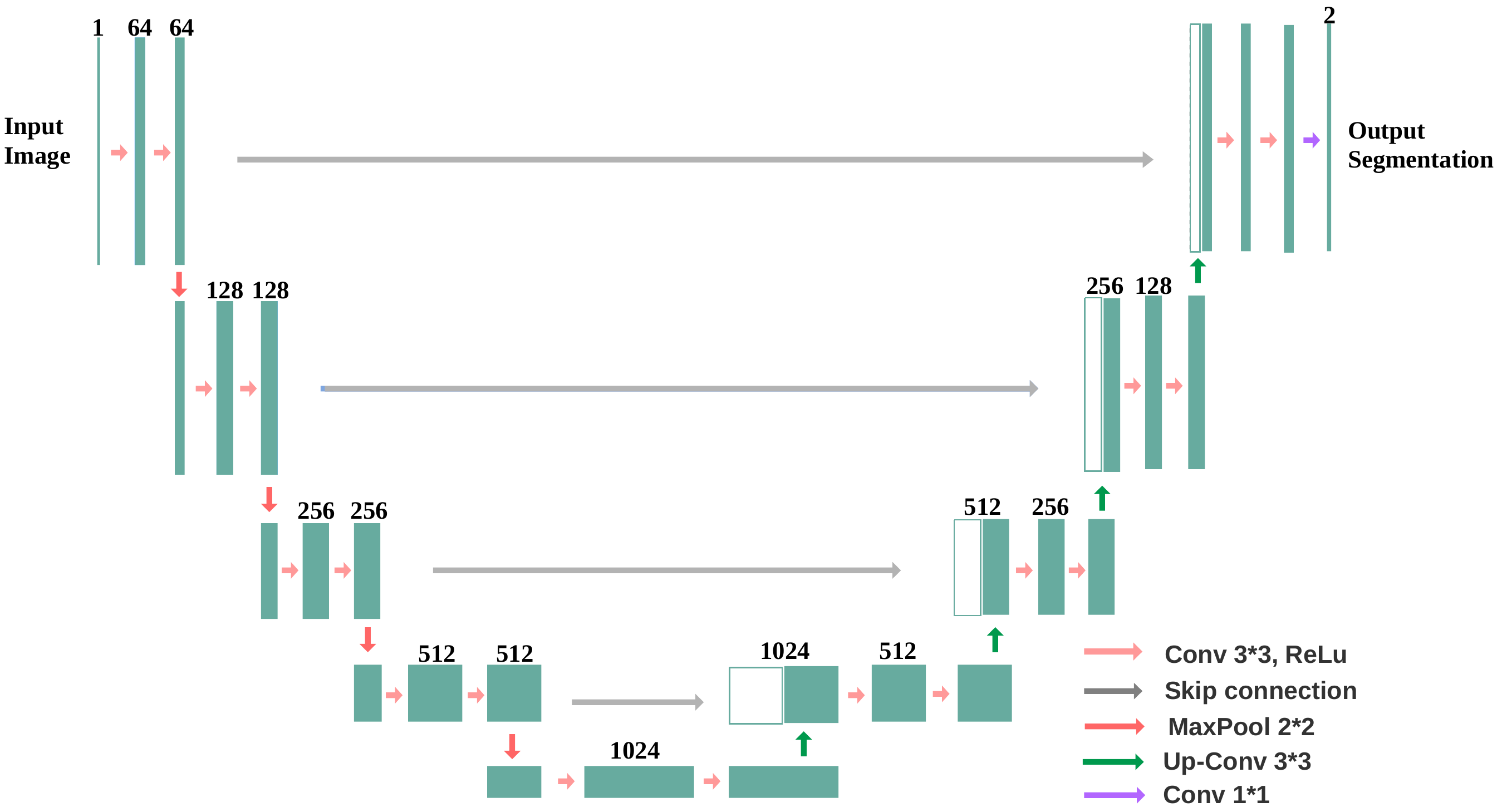}}
%    \subfloat[GAN ]{\includegraphics[width=0.49\textwidth]{U-net1.pdf}}
 %   \subfloat[a]{\includegraphics[width=0.48\textwidth]{StudentTeacher10.png}}
    \caption{U-Net in \cite{ronneberger2015u}. }
  \label{fig:02}
  \end{figure}

\textbf{Deep network for image segmentation.}  The convolutional neural network (CNN) has been the \textit{de-facto} solution for medical image segmentation. CNNs have shown striking improvement over traditional methods, such as machine learning methods using  hand-crafted features \cite{shotton2008semantic,peng2019mitochondria}, graph cut methods \cite{boykov2001interactive}, shape deformation \cite{cootes1995active}, and variational methods \cite{chan2006algorithms}. Image segmentation has recently been tackled by  end-to-end learning and fully convolutional networks (FCN) \cite{long2015fully}, especially in  encoder-decoder architecture \cite{noh2015learning,ronneberger2015u}. Compared to classical CNN, the FCN is composed of convolutional layers without any fully-connected layer at the end of the network and can  transform the  feature map of the intermediate layer back to the size of the input image. Thus, the prediction of an FCN has  a spatial one-to-one correspondence with the input image, which has dramatically  promoted semantic segmentation research. Many models with improved network architectures have been introduced, such as SegNet \cite{noh2015learning} with encoder-decoder architecture, the U-Net \cite{ronneberger2015u}, PSP-Net \cite{zhao2017pyramid} with pyramid pooling, DeepLab \cite{chen2017deeplab} with atrous spatial pyramid pooling, Attention U-Net with attention module\cite{oktay2018attention}, etc.
An FCN in encoder-decoder architecture typically consists  of  a contracting sub-net, i.e., the encoder,  that gradually reduces the feature maps and captures high-level features,
and  an expending sub-net, i.e., the decoder,  that gradually recovers the spatial information and fine boundary. A demonstration of the difference between a CNN and an FCN in encoder-decoder architecture  is shown in Fig. \ref{fig:01}.   Notably, the U-Net introduces  additional skip connections between the encoder and decoder (as shown in Fig. \ref{fig:02}), and  have produced very impressive results in the domain of medical image segmentation. Dense skip connections were  introduced in the DenseNet \cite{huang2017densely}, and have been widely used in many segmentation models \cite{litjens2017survey}.   Please refer to \cite{hao2020brief,taghanaki2021deep,litjens2017survey} for comprehensive reviews of recent improvements of the FCN models  for semantic segmentation of natural and medical images.

\textbf{Image segmentation with limited supervision.}  The cost of labor-intensive,
pixel-level annotation of large scale medical imaging data can be reduced by utilizing 1) a small subset of labeled training data, also known as semi-supervised learning or few-shot learning; 2) partial annotations (including sparse annotations), i.e., partially-supervised learning; or 3) inaccurate annotations including noisy labels, bounding boxes, and boundary scribbles. It is noteworthy that, though the labeled data is scarce in the semi-supervised setting, these annotated  are typically assumes to be   precise and reliable, which is different from inaccurate and partial annotation settings. The extreme case of limited supervision is the unsupervised setting, where there is ultimately no labeled data available. However, methods only exploring unlabeled data, such as clustering, are usually task-agnostic  and usually show very low performance for the complicated  segmentation tasks. Recently, auxiliary tasks, such as the adaptation of a well-trained model from a similar domain with a similar task \cite{kamnitsas2017unsupervised,peng2020unsupervised}, have been leveraged  to migrate this problem. Although we will not cover the unsupervised segmentation and their solutions, such as unsupervised domain adaptation (UDA) \cite{toldo2020unsupervised} and zero-shot learning \cite{wang2019survey}, we mention it here to start by looking at all settings in the big picture. In this paper, we focus on methods that learn to segment medical images with incomplete, inexact, and inaccurate annotations by jointly leveraging  a few labeled data and a large number of unlabeled examples.

  \section{Semi-supervised Segmentation} \label{sec:3}
Semi-supervised Segmentation  is a common scenario in medical applications, where only a small subset of the training images are assumed to  have full pixel-wise annotations. However, there is also an abundance of unlabeled images that can be used to  improve both the accuracy and generalization capabilities.  Since  unlabeled data do not involve labor-intensive annotations, any performance gain conferred by using unlabeled data  comes with a low cost. The major challenge of this learning scenario lies in how to efficiently and thoroughly exploit a large quantity of unlabeled data. The
most common approaches for semi-supervised segmentation include 1) general strategies, e.g.,  transfer learning, data augmentation, prior knowledge learning, curriculum learning, and few-shot learning, and 2) specialized methods that make use of unlabeled data, e.g.,  self-training \cite{bai2017semi,sedai2019uncertainty,yu2019uncertainty,li2020self}, consistency regularization \cite{bortsova2019semi},   co-training, self-supervised learning, and adversarial learning.

\begin{figure}[!t]
    \centering
    \includegraphics[width=0.49\textwidth]{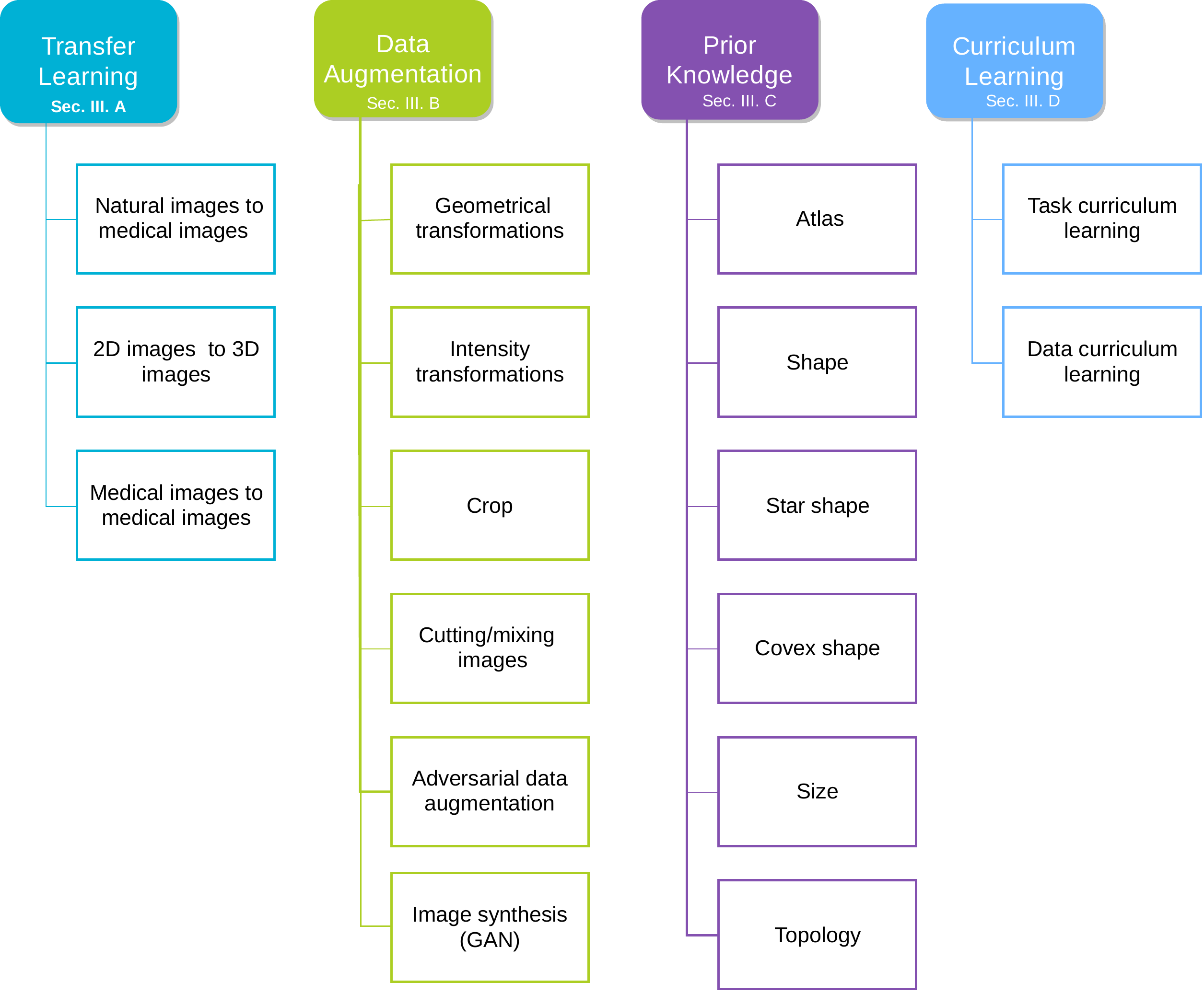}
    \caption{General solutions for tackling  semi-supervised medical image segmentation.     }
  \label{fig:2}
  \end{figure}

In the following subsections, we first review general methods that can be used to address small labeled data problem, i,e., transfer learning, data augmentation, prior knowledge learning, curriculum learning, as shown in Fig. \ref{fig:2}. Then, we discuss specialized methods designed for semi-supervised learning. Finally, we elaborate on  few-shot learning, which learns to generalize from a few examples with prior knowledge.
%  More recently, the
%attention has been shifted towards

 \subsection{Transfer learning}
Transfer learning refers to reusing a model developed for a task  as the starting point for a model on a second task, which may speed up the learning process, alleviate the problem of limited training data, and  improve generalization on the second task. In contrast, training an entire deep network from scratch usually requires  a large-scale labeled dataset. The "model pretraining and fine-tuning" strategy, a notable example of transfer learning, has been a simple but effective paradigm in many deep network applications since  many tasks are related. In many deep learning studies, transfer learning also narrowly  refers to the   "model pretraining and fine-tuning" strategy.
Typically, transfer learning from natural image data to medical datasets involves starting with standard network architectures, e.g., VGG \cite{simonyan2014very} and ResNet \cite{he2016deep},  and using  corresponding pre-trained weights trained on large-scale external sources of natural images, namely ImageNet \cite{deng2009imagenet} and PASCAL VOC \cite{everingham2010pascal},  as initialization or a fixed feature extractor,  and then fine-tuning the model on  medical imaging data.  This model reusing strategy tends to work if the features are general and suitable for both the source and target tasks. The transferability of features on different layers of the deep network was investigated in \cite{yosinski2014transferable}. They showed that transferring features even from distant tasks can be better than using random features.

  It is noteworthy that, various medical applications involves segmentation of 3D medical  images, which hinders the transfer of pre-trained model on 2D natural images to the current task. While it is straightforward  to reformulate volume image segmentation to slice-by-slice 2D segmentation, rich spatial 3D contexts will inevitably lose.
  Possible solutions for transferring  2D networks to 3D networks include  1)  copying the 2D kernels along an axis \cite{carreira2017quo} and 2) padding the pre-trained 2D kernels by zeros along an axis \cite{shan20183,liu20183d}. For instance, Liu \textit{et al.} \cite{liu20183d} proposed to  transfer convolutional features learned from 2D
images to 3D anisotropic volume and obtained  desired strong generalization capability of the pre-trained 2D network.

For medical image analysis, another challenge of the pre-trained model on large-scale natural image sets is the significant domain gap between natural images and medical images, even an obvious gap between medical images of different modalities.  Tajbakhsh \textit{et al.} \cite{tajbakhsh2016convolutional} investigated the effectiveness of pre-trained deep CNNs with sufficient fine-tuning compared to training a deep network from scratch on  four different medical imaging applications. They showed that, in most cases, fine-tuned network could outperform those trained from scratch and showed better robustness. However, Raghu \textit{et al.} \cite{raghu2019transfusion} recently evaluated the properties of transfer learning from ImageNet on two large scale medical imaging tasks. They demonstrated a contrasting result that  transfer learning gained little performance benefit, and simple and lightweight models can perform comparably to large pre-trained networks. Zoph \textit{et al.} \cite{zoph2020rethinking} demonstrated that stronger data augmentation and more labeled data would diminish the benefit of pretraining for vision applications, but self-training is always helpful.

While a well pre-trained model trained on a large-scale medical image dataset may be more valuable for medical image segmentation, there is not a large scale  annotated dataset like
ImageNet in the medical domain.  To obtain a universal  pre-trained model with promising transferable and generalizable ability for medical image analysis, several studies have proposed to pre-train models on medical datasets that are limited to specific modalities or tasks. Zhou \textit{et al.} \cite{zhou2019models}  built a 3D pre-trained model, called Genesis Chest CT, using unlabeled 3D Chest Computed Tomography (CT) images with a novel self-supervised learning method. Similar pre-trained models for specific image domains were also similarly built, such as  Genesis Chest CT 2D, and Genesis Chest
X-ray, which used 2D chest CT and chest X-ray images, respectively. A  universal 3D model was learned in \cite{zhang2020universal} by leveraging
a self-supervised learning scheme from multiple unlabeled source datasets of different modalities and distinctive scan regions.

\subsection{Data augmentation} Since deep networks are heavily reliant on big data to learn discriminative representation  and  avoid overfitting, data augmentation \cite{lecun1998gradient} has been  considered as a simple yet effective data-space solution to the problem of limited annotated data. Specifically, data augmentation  aims to artificially enhance the size, diversity, and quality of the training data without collecting and manually labeling new data.
Typical data augmentation methods not only include  data warping methods \cite{lecun1998gradient} such as random affine and  elastic transformations, random cropping \cite{he2016deep}, random erasing \cite{zhong2017random,devries2017improved}, intensity transformation and adversarial data augmentation \cite{volpi2018generalizing,madry2018towards}, but also include methods that can synthesize more diverse and realistic labeled examples, such as mixing images \cite{zhang2017mixup,yun2019cutmix,guo2019mixup,panfilov2019improving},  feature space augmentation \cite{devries2017dataset}, and generative adversarial networks \cite{goodfellow2014generative,yi2019generative,tegang2020gaussian,valanarasu2020learning}.  While general transformation augmentation methods such as random affine transformations, elastic transformations, and intensity transformations are  easy to implement and have shown performance improvements in abundant applications \cite{ronneberger2015u,yuan2020net,luo2020hdc}, they do not take advantage of the knowledge in unlabeled training data. Recently, there is a growing interest in developing augmentation that can simulate real variations of the data, and thus task-driven approach \cite{zhao2019data,chen2020realistic,oksuz2019automatic} is a promising direction. Schlesinger \textit{et al.} \cite{schlesinger2020deep} provided a recent review of data augmentation methods for  brain tumor segmentation.

\textbf{Mixing and cutting images} \cite{zhang2017mixup,yun2019cutmix,guo2019mixup,verma2019manifold} is a class of simple but effective augmentation methods in many applications \cite{panfilov2019improving}. Specifically, Mixup \cite{zhang2017mixup} linearly interpolates a random
pair of training images and correspondingly their labels. Recently, Mixup has been improved in \cite{guo2019mixup} with learned mixing policies to prevent manifold intrusion. Cutout \cite{devries2017improved} adopts the idea of the regional dropout strategy, that is occluding a portion of an image, on training data.  Alternatively, CutMix \cite{yun2019cutmix} is a combination of aspects
 of Mixup and Cutout by replacing a portion of an image with a portion of a different image.
For the application of medical image segmentation, Panfilov \textit{et al.} \cite{panfilov2019improving} tested the efficiency of Mixup for knee MRI segmentation and showed improved model robustness.

\textbf{Adversarial data augmentation }involves harnessing adversarial examples to train robust models against unforeseen data corruptions or distribution shifts \cite{goodfellow2014explaining,madry2018towards,volpi2018generalizing} and thus is  plausible  to cope
with limited labeled training data.  When applying to medical image segmentation, designing and constructing more realistic adversarial perturbations is a crucial problem \cite{ratner2017learning,volpi2018generalizing,chen2020realistic}. For  MR image segmentation,  Chen \textit{et al.} \cite{chen2020realistic} introduced intensity inhomogeneity as a new type of adversarial attack  using a realistic intensity transformation function learned with adversarial training  to amplify intensity non-uniformity in MR images  and simulate potential image artifacts, such as bias field. To obtain
adversarial samples subject to a given transformation model, Olut \textit{et al.} \cite{olut2020adversarial} proposed to learn a statistical deformation model that can capture plausible anatomical variations from unlabeled data  by deep registration models. A similar idea was adopted in \cite{shen2020anatomical}.

\textbf{Generative adversarial networks (GAN)} \cite{goodfellow2014generative,yi2019generative} have also been utilized to conduct medical data augmentation by  directly synthesizing new labeled data.  Costa \textit{et al.} \cite{costa2017end} proposed training   a generative model  with adversarial learning to synthesize both realistic retinal vessel trees  and retinal color images.   For semi-supervised medical image segmentation, Chaitanya \textit{et al.} \cite{chaitanya2019semi} proposed  to learn a generative network  to synthesize new samples from both labeled and unlabeled data by simultaneously learning and applying realistic spatial deformation fields and additive intensity transformation fields.  To improve cross-modal  segmentation with limited training samples,  Cai\textit{ et al.} \cite{cai2019towards} developed a cross-modality data synthesis approach to generate realistic looking 2D/3D images of a specific modality as data augmentation. Yu \textit{et al.} \cite{yu2019ea} integrated edge information into conditional GAN \cite{mirza2014conditional} for  cross-modality MR image synthesis. To segment pulmonary nodules, Qin \textit{et al.} \cite{qin2019pulmonary}  augmented the training set with synthetic  CT images and labels and achieved promising results.   For one-shot brain segmentation, Zhao \textit{et al.} \cite{zhao2019data} used a data-driven approach for synthesizing labeled  images as data augmentation. Specifically, they  proposed to model the set of spatial and appearance transformations between all the training data, including both the labeled and unlabeled images, and then applied the learned transformations on the single labeled image to synthesize new labeled images.

\subsection{Prior knowledge learning }
A group of semi-supervised methods has addressed semi-supervised segmentation by incorporating  prior knowledge/ domain knowledge  such as anatomical priors about the objects of interest  into the segmentation model as a strong regularization \cite{wang2020lt,xu2019deepatlas,el2020bb,oktay2017anatomically,dalca2018anatomical,larrazabal2019anatomical}. In fact, prior knowledge about location, shape, anatomy, and context is also crucial for manual annotation, especially
in the presence of fuzzy boundaries or low image contrast. As for semantic segmentation with deep networks,   the model training is typically guided by local or
pixel-wise loss functions (e.g., Dice loss \cite{milletari2016v} and cross-entropy loss), which may not be sufficient
to learn informative features about the underlying anatomical
structures and global dependencies. Anatomical-prior guided methods usually assume  the plausible solution space can be expressed in the form of a prior distribution, enforcing  the network to generate more
anatomically plausible segmentations.

 \textbf{Atlas-based segmentation} \cite{iglesias2015multi,cabezas2011review} with single- or multiple- atlas has been widely used in medical image segmentation to exploit prior  knowledge from previously labeled training images. An atlas consists of a reference model with labels related to the anatomical structures. Thus, it can provide crucial knowledge, such as  information about location, texture, shape, spatial relationship, etc., for segmentation, especially when limited labeled data available for model training. Atlas-based methods essentially treat the segmentation problem as a registration problem, and  non-rigid registration is typically used  to account for the anatomical differences between subjects.  Wang \textit{et al.} \cite{wang2020lt} addressed one-shot segmentation of brain structures from Magnetic Resonance Images (MRIs) with single atlas-based segmentation, where reversible voxel-wise correspondences between the atlas and the unlabelled images were learned with a correspondence-learning deep network.
 Ito \textit{et al.} \cite{ito2019semi} considered semi-supervised segmentation of  brain tissue from MRI. Specifically, they relied on image registration with one or more atlas to generate pseudo labels on unlabeled data. The  expectation-maximization (EM) algorithm was used to update model parameters and pseudo labels alternatively. However, image registration,  the process of geometrically aligning two or more images,  is computation-intensive, which may hamper its practical application. Please refer to \cite{de2019deep} for a comprehensive review of both affine and deformable image registration with deep learning methods.  A similar idea was employed by Chi \textit{et al.} \cite{chi2020deep}, where they generated pseudo-labels by utilizing deformable image registration to propagate atlas labels onto  unlabeled images.  Xu and Niethammer \textit{et al.} \cite{xu2019deepatlas} proposed to jointly learn two deep  networks for weakly-supervised image registration and semi-supervised segmentation,  assuming that these two tasks can mutually guide each other's training on unlabeled images. He \textit{et al.} \cite{he2019dpa} further proposed an improved joint learning model, which added a perturbation factor in the registration to its sustainable data augmentation ability and a discriminator to extract registration confidence maps for better guidance of the segmentation task. For 3D left cavity (LV) segmentation on echocardiography with limited annotation data,  Dong \textit{et al.} \cite{dong2020deep} introduced a deep atlas network with a lightweight registration network and a multi-level information consistency constraint. However, registration, which is a computation insensitive and challenging task, is not an essential segmentation. For semi-supervised 3D liver segmentation, Zheng \textit{et al.} \cite{zheng2019semi} proposed to combine probabilistic atlas, which can provide the shape and position prior,  with deep segmentation  networks using a prior weighted cross-entropy loss. The probabilistic atlas was obtained by averaging the manually labeled liver masks after aligning all labeled training images.   Vakalopoulou \textit{et al.} \cite{vakalopoulou2018atlasnet} developed an AtlasNet, which consisted of multiple deep  networks trained after co-aligning multiple anatomies through multi-metric deformable registration. The multiple deep networks were used to map all training images to common subspaces to reduce biological variability.

 \textbf{Shape-prior based  segmentation}  \cite{oktay2017anatomically,dalca2018anatomical,larrazabal2019anatomical,larrazabal2020post,mirikharaji2018star,duan2019automatic,lee2019tetris,lucontour,tilborghs2020shape,yue2019cardiac}
  has been an active research topic in the context of deep learning to obtain more accurate
and anatomically plausible segmentation. While principal component analysis (PCA) based
statistical shape model (SSM) \cite{cootes1995active} was widely adopted by  traditional segmentation methods, it is not straightforward to combine SSM with deep networks. Ambellan \textit{et al.} \cite{ambellan2019automated} combined 3D SSMs with 2D
and 3D deep convolutional networks to obtain a robust and accurate segmentation of even highly pathological knee bone and cartilage. Specifically, they used SSM adjustment as a shape regularization of the outputs of the segmentation networks.  Oktay \textit{et al.} \cite{oktay2017anatomically} initially used a stacked convolutional autoencoder  to learn  non-linear shape representations, which is integrated with the segmentation network to enforce its predictions  to follow the learned anatomical priors. With the shape prior, their method obtained highly competitive
performance for  cardiac image segmentation while learning from a limited number (30) of labeled cases.  Rather than using the compact codes  produced by an autoencoder as the shape constraint in \cite{oktay2017anatomically},  Yue \textit{et al.} \cite{yue2019cardiac} used the reconstructions of the predicted segmentations to maintain a realistic shape of the resulting segmentation.

While the framework in \cite{oktay2017anatomically,yue2019cardiac} incorporated the learned anatomical prior into deep networks through a regularization  term, Painchaud \textit{et al.} \cite{painchaud2020cardiac} incorporated the anatomical priors through an additional post-processing stage. Specifically, they warped initial segmentation results toward the closest anatomically correct
cardiac shape, which was leaned  and generated with a constrained variational autoencoder. Ravishankar \textit{et al.} \cite{ravishankar2017learning} introduced a shape regularization network (convolutional autoencoder) after the segmentation. Larrazabal \textit{et al.} \cite{larrazabal2020post} learned lower-dimensional representations of plausible shapes with a denoising autoencoder and use it as a post-processing step to impose shape constraints on the coarse output of the segmentation network.

As a novel extension of template deformation methods \cite{cootes1995active} in the context of deep networks, Lee \textit{et al.} \cite{lee2019tetris} introduced a template transformer
network, where a shape template is deformed to match the underlying structure of interest through an end-to-end
trained spatial transformer network.  Zotti \textit{et al.} \cite{zotti2018convolutional} introduced a probabilistic image  estimated by computing the pixel-wise empirical proportion of each class based on aligned  ground truth label fields of the training images. The probabilistic shape-prior image was concatenated with network features for prior guidance. For the semi-supervised 3D segmentation of renal artery, He \textit{et al.} \cite{he2019dpa} proposed assisting the segmentation network with multi-scale semantic features extracted from unlabeled data with an autoencoder.

Other types of anatomical priors such as star shape prior \cite{mirikharaji2018star,liu2020deep,weigert2020star,rak2019combining,schmidt2018cell}, convex shape prior \cite{liu2020convex}, topology \cite{bentaieb2016topology,huang2018medical,clough2019explicit,clough2020topological,byrne2020persistent}, size \cite{kervadec2019constrained,bateson2019constrained,chen2019learning}, etc., have also been introduced to improve the segmentation robustness and anatomically accuracy.

    \begin{figure*}[!t]
    \centering
     \subfloat[Self-training]{\includegraphics[width=0.4\textwidth]{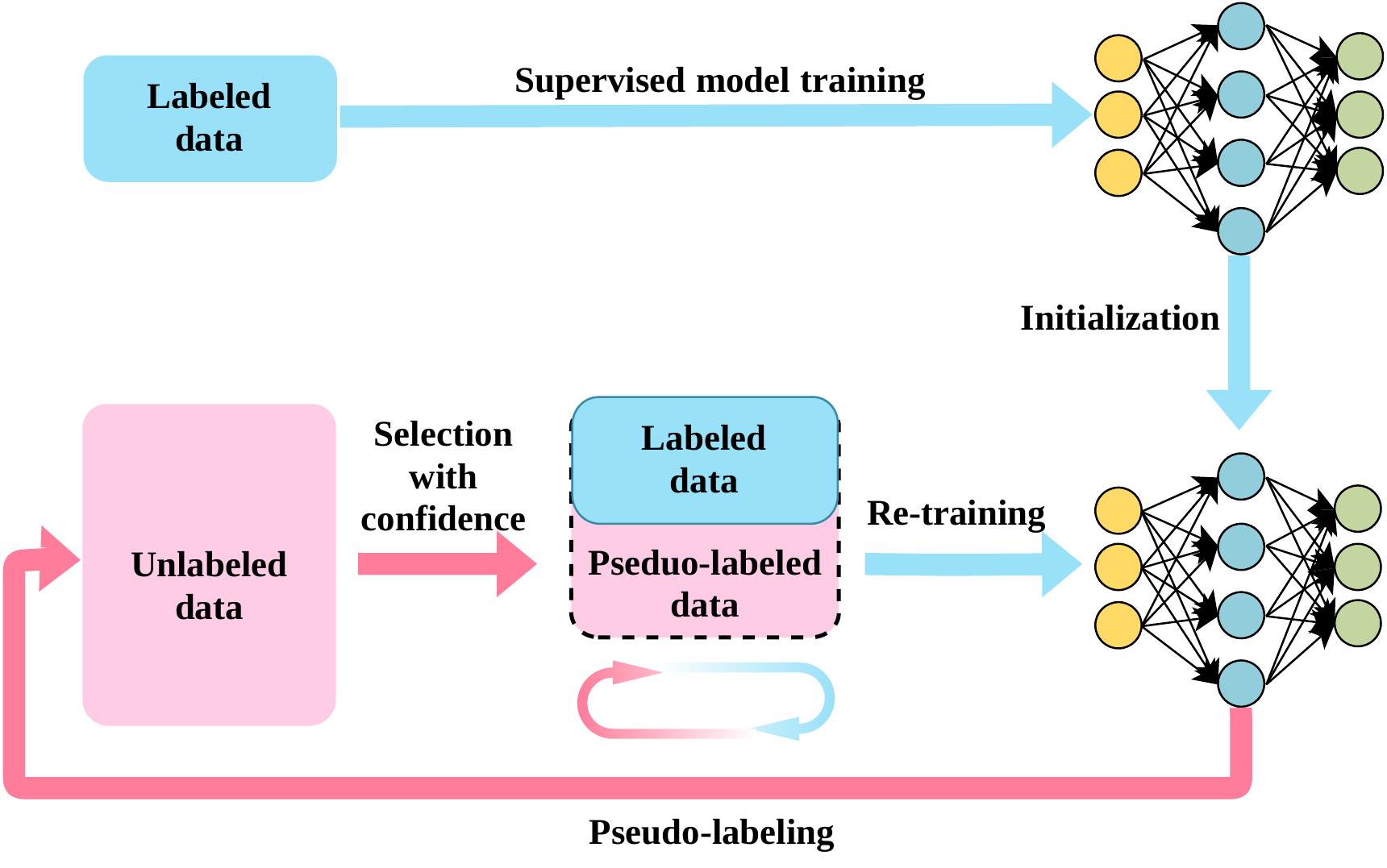}}\hspace{5mm}
      \subfloat[Active learning]{\includegraphics[width=0.4\textwidth]{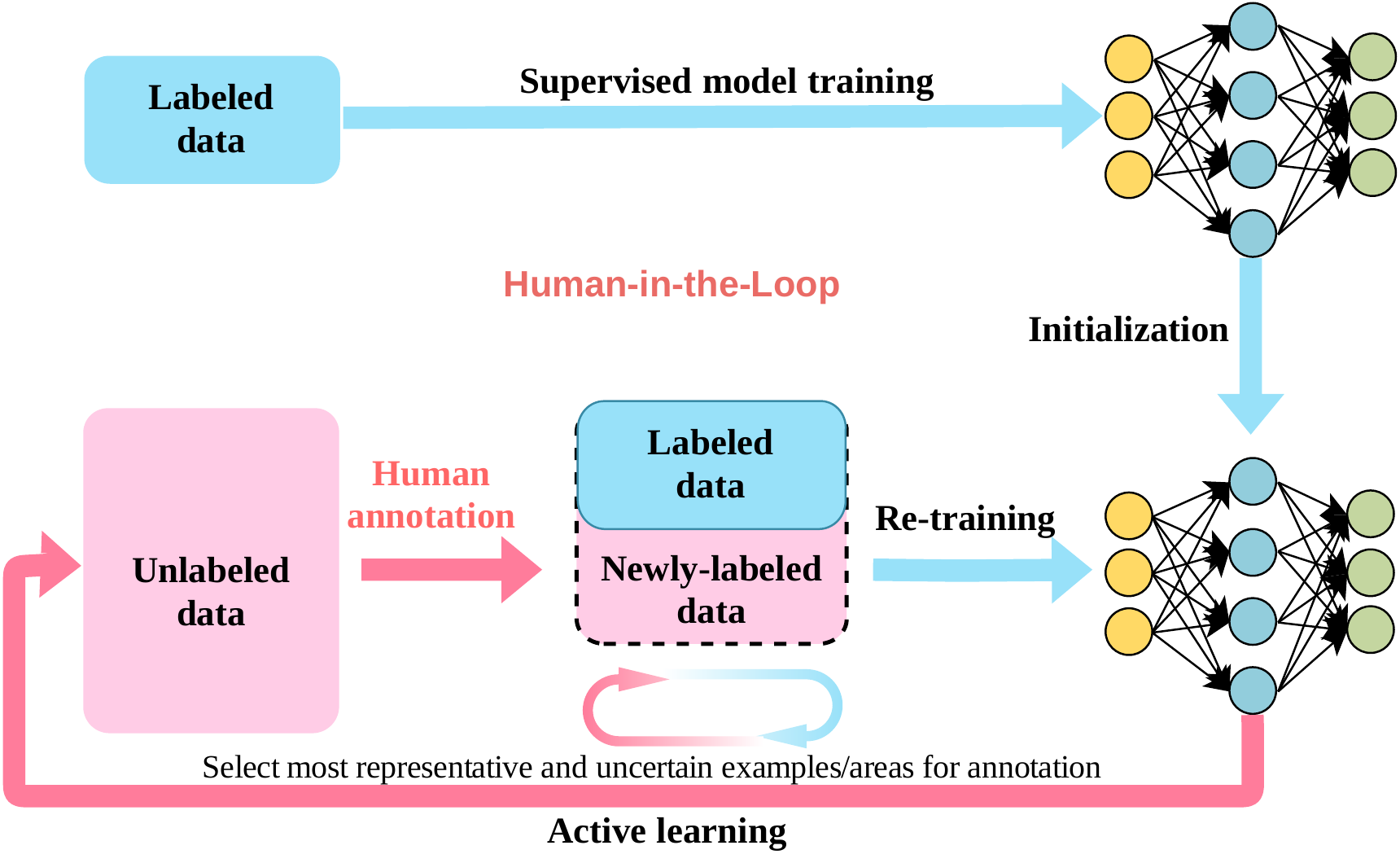}}%
    \caption{ The flowchart of the self-training for semi-supervised segmentation  and active learning  for interactive segmentation on a conceptual level. For self-training, the segmentation model is initially trained only on the small size labeled data and then retrained with the extended  data that consists of both the labeled data and confidence pseudo-labeled data. Human interaction is required for active learning, and the segmentation model is retrained with original labeled data and newly-labeled data.  }
  \label{fig:3}
  \end{figure*}

\subsection{Curriculum learning} Given the greater complexity of semi-supervised segmentation over classification and \textit{the importance of starting small}, the concept of curriculum learning \cite{bengio2009curriculum}, or the easy-to-hard strategy, has also been utilized. Curriculum learning describes a type of learning strategy that  first starts with  easier aspects of the task or easier subtasks and then gradually increases the difficulty level.  In a broad sense, the most widely used  curriculum learning strategies  include \textit{data curriculum learning} and  \textit{task curriculum learning}. While early studies  focused on data curriculum learning by reweighting the target training distribution, recent studies also investigated  the varied easiness among different works \cite{pentina2015curriculum}, i.e., task curriculum learning.

 In data  curriculum learning, non-uniform sampling of examples or mini-batches from the entire training data, rather than uniform-sampling as in standard deep network training, is used in  model training. Therefore, the core tasks are how to rank the training examples and how to  guide the order of presentation of
examples based on this ranking \cite{hacohen2019power}. Thus, it is flexible to incorporate prior knowledge about the data and task. It has been empirically demonstrated that this learning paradigm is
useful in  avoiding bad local minima and in achieving better generalization ability \cite{khan2011humans}.
 Data  curriculum learning has recently been used in several medical applications, especially location and classification tasks \cite{tang2018attention,zhao2020egdcl,jimenez2019medical,oksuz2019automatic} but few in segmentation tasks \cite{jesson2017cased,wang2018deep}.   To train a deep network for the classification and  location of thoracic diseases on chest
radiographs, Tang \textit{et al.} \cite{tang2018attention} first ranked the  training images  according to the difficulty (indicated by the severity-levels of the disease) and then fed them to the deep network to boost the representation learning gradually. For fracture classification, Jim{\'e}nez-S{\'a}nchez \textit{et al.} \cite{jimenez2019medical,jimenez2020curriculum} assigned a degree of difficulty to each training example according to medical decision trees and inconsistencies in multiple experts' annotations.  In addition to the predefined curriculum   by prior knowledge and keeping it  fixed after that, the curriculum can also be dynamically determined
to adapt to the feedback of the learner, also known as \textit{self-paced curriculum learning} \cite{jiang2015self}  or \textit{self-paced learning} \cite{kumar2010self}. For lung  nodule segmentation/detection with extreme class imbalance,  Jesson \textit{et al.} \cite{jesson2017cased} introduced an adaptive sampling strategy, which  favors difficult-to-classify examples. For instance-level segmentation of pulmonary nodule,  Wang \textit{et al.} \cite{wang2018deep} employed pseudo labels as the surrogate of ground truth labels on unlabeled data. To  utilize the pseudo-labeled data, they followed the idea of self-spaced curriculum learning \cite{jiang2015self} and embedded curriculum design as a
regularization term into the learning objective.

 Task curriculum learning consists of tackling easy but related tasks first  to provide auxiliary  information for more complicated tasks, which will be solved later. Task curriculum learning is highly related to multi-stage learning in segmentation \cite{dai2016instance,vigneault2018omega,luo2020hdc}, where more easier tasks such as location or  coarse segmentation are first solved with a simple method. After that,  the more complex pixel-level segmentation is addressed. For example, for cross-domain segmentation of natural images, Zhang \textit{et al.} \cite{zhang2017curriculum} proposed to  solve easy tasks first to infer necessary properties about the target domain. Specifically, they first estimated label distributions over both global images and
some landmark superpixels of the target domain. They then enforced the semantic segmentation network to follow those target-domain properties as much as possible.
For left ventricle segmentation in MR images, Kervadec \textit{et al.} \cite{kervadec2019curriculum} introduced a curriculum-style strategy  that first learned  the size of the target
region, which is the more easier task and then regularized the segmentation task, which is a difficult task, with pre-learned region size.

\subsection{Self-training strategy }\label{subsec: selftraining}
 Self-training \cite{lee2013pseudo,triguero2015self,bai2017semi,li2019signet} (also called pseudo-labeling) is an iterative process that alternatively generates pseudo-labels on the unlabeled data and  retrains  the learner  on the combined labeled and
pseudo-labeled data. An illustration of the self-training based semi-supervised segmentation is shown in Fig. \ref{fig:3} (a) (with a comparison of the strategy of human-in-the-loop, i.e., active learning for interactive segmentation that will be discussed in Sec. \ref{sec:4}).  The   generality and flexibility of  self-training have been validated in many  applications \cite{zoph2020rethinking}. A fundamental property of self-training strategy is that it can be combined with any supervised learner and provides  a straightforward but effective manner of leveraging unlabeled data.

\begin{figure}[!t]
    \centering
    \includegraphics[width=0.4\textwidth]{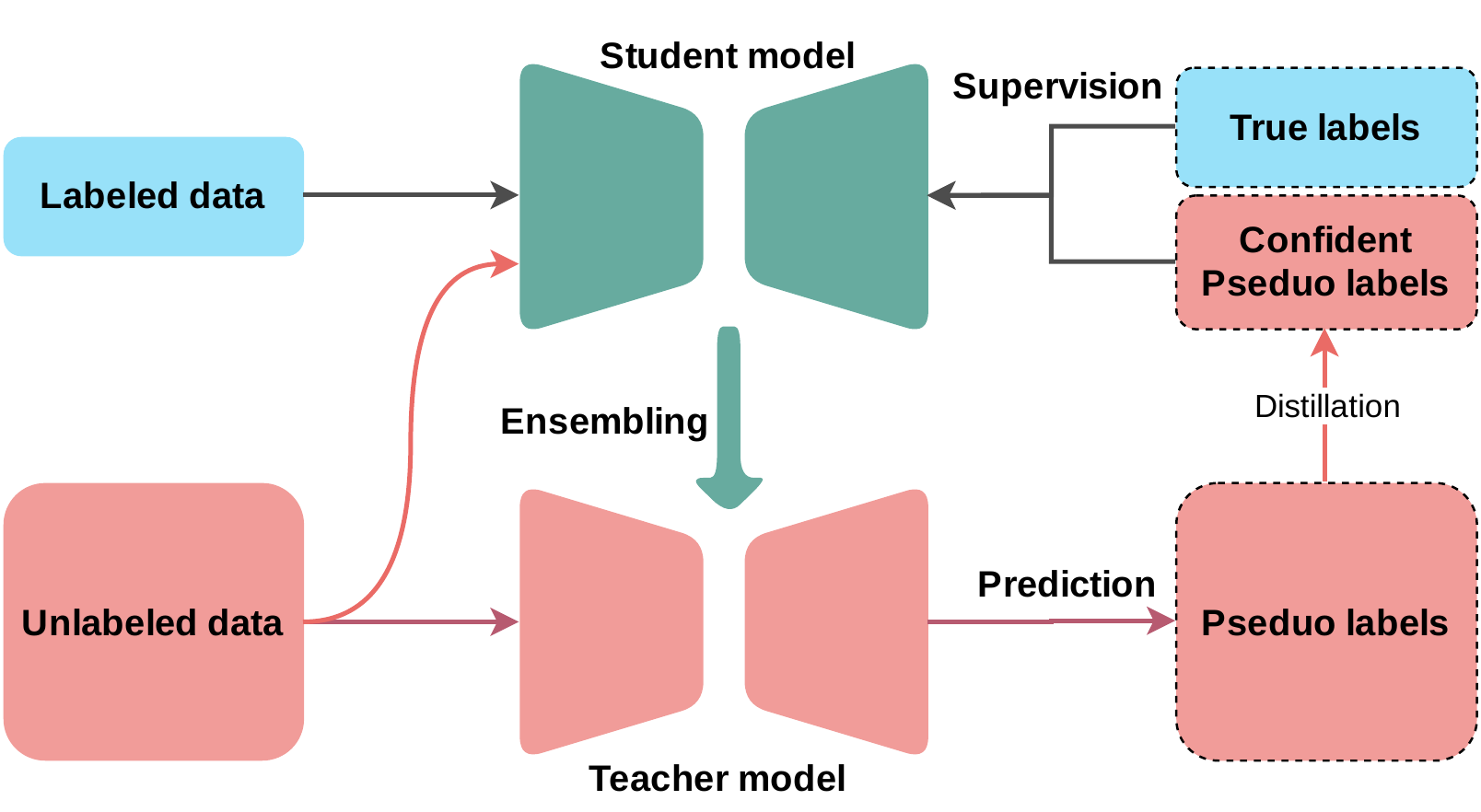}%
    \caption{ The flowchart of a typical self-training procedure in the teacher-student framework on a conceptual level.  The segmentation model is initially trained only on the small size labeled data and then retrained with the extended training dataset that consists of both the labeled data and confident pseudo-labeled data.  }
  \label{fig:4}
  \end{figure}

  Self-training is usually in a teacher-student paradigm \cite{laine2016temporal,tarvainen2017mean} (as shown in Fig. \ref{fig:4}), which consists of first learning a teacher model from ground truth annotations and then using the predictions of the teacher model to generate pseudo-labels on the unlabeled data.  The ground truth annotations and pseudo labels with high confidence are further jointly digested  iteratively  to learn a powerful student model.  Bai \textit{et al.} \cite{bai2017semi} first trained a teacher neural network using the labeled data and then utilized prediction confidence (i.e., the probability prediction followed  by a
conditional random field (CRF) refitment) of the teacher model on the unlabeled data as the pseudo labels. Fan \textit{et al.} \cite{fan2020inf} applied a similar strategy for lung infection segmentation from CT images. Typically, the self-training method is iterative, and the quality of pseudo-labels should be  gradually improved for a successful self-training approach \cite{tarvainen2017mean}. The self-training strategy's main challenge  lies in generating reliable pseudo labels and handling the negative impacts by adding incomplete and incorrect  pseudo-labels, which may  confuse the model training.

 A promising direction to improve the quality of pseudo labels and reduce the negative impact of noisy pseudo labels is to estimate  the uncertainty or confidence estimation \cite{gal2016dropout,der2009aleatory,blundell2015weight,ovadia2019can,kendall2017uncertainties}.  To this end,  it is pleasable to let the teacher model simultaneously  generate the segmentation predictions as the pseudo labels and estimate the uncertainty maps for the unlabeled images. The uncertainty maps can be used as  guidance to
maintain  reliable predictions \cite{nair2020exploring,graham2019mild,jungo2019assessing,cao2020uncertainty,wen2019bayesian,mehrtash2020confidence,wang2019aleatoric,hiasa2019automated}.

There are two  categories
of uncertainty  \cite{der2009aleatory,kendall2017uncertainties}  one can model, namely \textit{ aleatoric uncertainty} (data uncertainty), which is an inherent property of the data
distribution and irreducible, and \textit{epistemic
uncertainties} (model uncertainty), which can be  reduced through the collection of additional
data.   Popular  approaches to generate pseudo-labels and quantify uncertainties in deep networks include Bayesian neural networks \cite{blundell2015weight,wang2016towards}, Monte Carlo Dropout \cite{gal2016dropout,yu2019uncertainty,sedai2019uncertainty,venturini2020uncertainty},  Monte Carlo batch normalization \cite{teye2018bayesian},  and  deep ensembles \cite{lakshminarayanan2017simple}.   Bayesian neural  networks capture model uncertainty by learning a posterior distribution over parameters. While Bayesian networks are often hard
to implement and computationally slow to train \cite{gal2016dropout,lakshminarayanan2017simple}, non-Bayesian strategies, including Monte Carlo Dropout and deep ensembles,  are more attractive. Jungo \textit{et al.} \cite{jungo2019assessing} evaluated several widely-used pixel-wise uncertainty measures concerning their reliability and limitations for  medical image segmentation, and also highlighted the
importance of developing subject-wise uncertainty estimations.

The model uncertainty estimated with Monte Carlo Dropout \cite{gal2016dropout} can be interpreted as  an approximation of Bayesian uncertainty. Concretely, the predictive uncertainty is estimated by averaging the results of multiple stochastic forward passes of the deep network under random dropout.
 The widely-used uncertainty measures include normalized entropy of the varied  probabilistic predictions, the variance of the Monte Carlo samples, mutual information, and predicted variance. Nair \textit{et al.} \cite{nair2020exploring} provided an in-depth analysis of the different measures based on medical image segmentation performance. Camarasa \textit{et al.} \cite{camarasa2020quantitative} conducted a quantitative and statistical comparison of several uncertainty measures of Monte Carlo Dropout  based on the task of multiclass segmentation.
 Given the uncertainty estimated by Monte Carlo Dropout \cite{gal2016dropout}, Yu \textit{et al.} \cite{yu2019uncertainty} introduced an uncertainty-aware consistency loss for the learning of the student model and applied it to the semi-supervised segmentation of the left atrium. Similarly, Sedai \textit{et al.} \cite{sedai2019uncertainty} conducted semi-supervised segmentation of  retinal layers in OCT images  with uncertainty guidance estimated with Monte Carlo Dropout.

 Deep ensembles \cite{lakshminarayanan2017simple} were  theoretically motivated by the bootstrap and have been empirically demonstrated to be a promising approach for boosting the accuracy and robustness of deep networks. Concretely, multiple networks using different training subsets and/or different initializations are  separately trained to enforce variability, and then the predictions are combined by averaging for the uncertainty estimation. Mehrtash \textit{et al.} \cite{mehrtash2020confidence} used  deep ensembles  for  confidence calibration, where they trained multiple models with different initializations and random shuffling of the training data. They applied their confidence calibrated model for  brain, heart, and  prostate segmentation.

  Ayhan and Berens \cite{ayhan2018test} showed that applying traditional augmentation at the test time can be an effective and efficient  estimation of heteroscedastic aleatoric uncertainty in deep networks, and they applied their method on  fundus image analysis.  Kendall and Gal \cite{kendall2017uncertainties} introduced a unified Bayesian framework to combine aleatoric and epistemic uncertainty estimations for deep networks. Wang \textit{et al.} \cite{wang2019aleatoric} validated the effectiveness of test-time augmentation as aleatoric uncertainty estimation on the segmentation of fetal brains and brain tumors.

      \begin{figure*}[!t]
    \centering
      \subfloat[Consistency learning in single model ]{\includegraphics[width=0.4\textwidth]{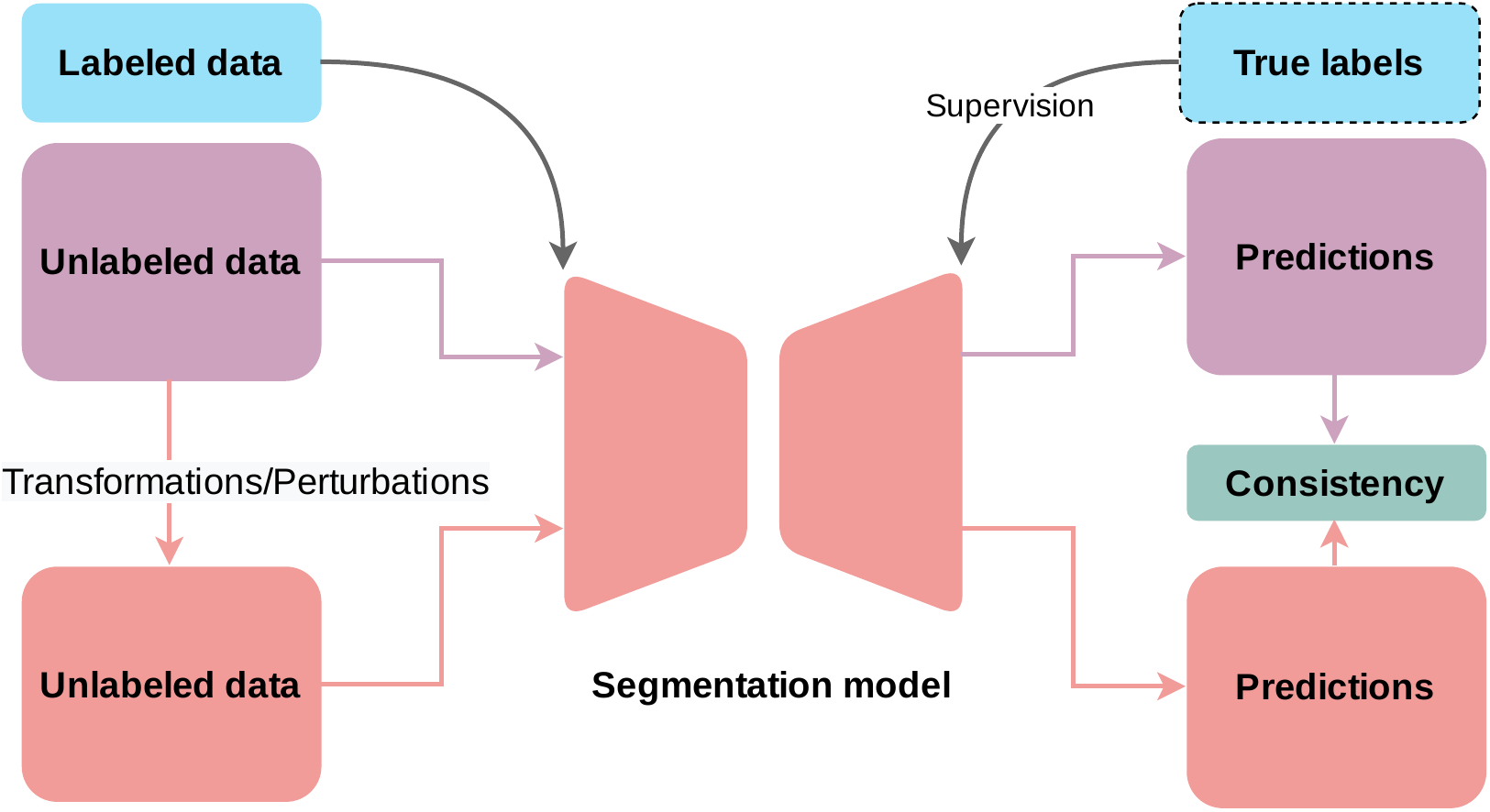}} \hspace{5mm}
    \subfloat[Consistency learning in dual model ]{\includegraphics[width=0.4\textwidth]{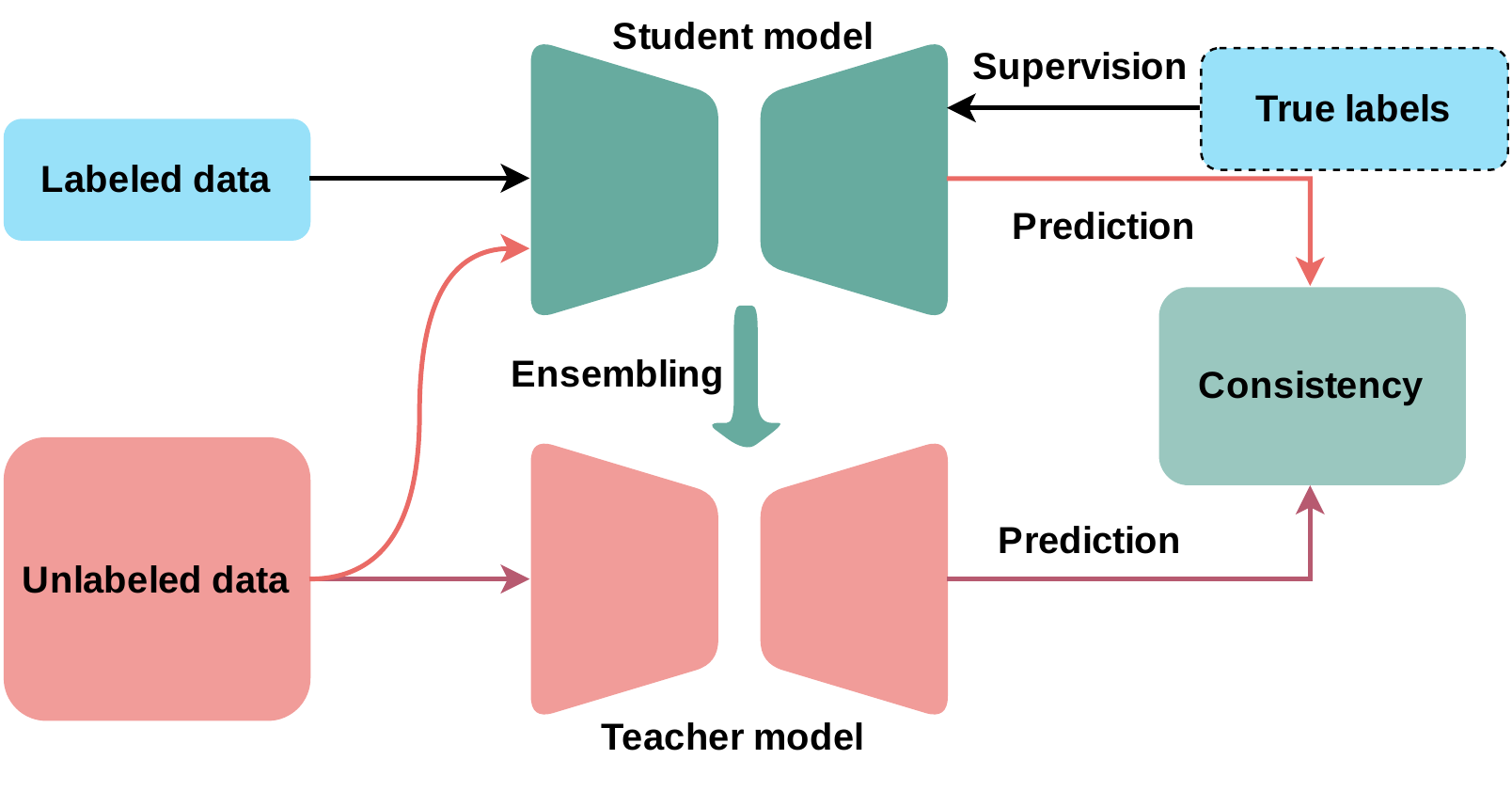}}
 %   \subfloat[a]{\includegraphics[width=0.48\textwidth]{StudentTeacher10.png}}
    \caption{ The conceptual idea of consistency regularization. a) Single model with consistency constraint, e.g., $\Pi$ model \cite{laine2016temporal}; b) dual model in the teacher-student framework with consistency constraint, e.g., Temporal Ensembling \cite{laine2016temporal}, and Mean Teacher \cite{tarvainen2017mean}. }
  \label{fig:5}
  \end{figure*}
 \subsection{Co-training}
Co-training initially introduced by Blum and Mitchell \cite{blum1998combining} exploits multiview data descriptions to learn from  a limited number of labeled examples and a large amount of unlabeled data. The underlying assumption is that the training examples can  be described by two or more different but
complementary sets of features, called views, which are assumed  conditionally independent  in the ideal given the category. As an extension of self-training to multiple base learners, the  original  co-training for classification  first learns a separate learner for each view using any labeled examples, then most confident predictions of all base learners on unlabeled data are gradually added to the labeled data of other base learners to continue the iterative
training.  By enforcing  prediction agreements between the different but related views, the goal is to  allow inexpensive unlabeled data to augment a much smaller set of labeled examples. Moreover, it is essential to ensure the different base learners giving different and complementary information about each instance \cite{wang2010new}, namely, \textit{view difference constraint} or \textit{diversity
criterion}. Peng \textit{et al.} \cite{peng2020deep} applied the idea of co-training to semi-supervised  segmentation of medical images. Concretely, they trained
multiple models on different subsets of the labeled training data and used a common set of unlabeled training images to exchange
information with each other.  Diversity across models was
enforced by utilizing adversarial samples generated using both the labeled and unlabeled data as \cite{qiao2018deep}.  For semi-segmentation of multi-organ from 3D medical images, Zhou \textit{et al.} \cite{zhou2018semi} introduced multi-planar co-training, which involves  training  different segmentation models on multiple planes, i.e., axial,
coronal, and sagittal planes, of a volume image in the teacher-student paradigm. Xia \textit{et al.} \cite{xia2020uncertainty} incorporated  uncertainty estimation to the multi-planar co-training approach in \cite{zhou2018semi} to generate more reliable pseudo labels for unlabeled data.

 \subsection{Consistency regularization } \label{subsec: Consistency}
 Consistency regularization \cite{sajjadi2016regularization,bachman2014learning,bortsova2019semi} utilizes unlabeled data by relying on the assumption that favoring models should generate consistent
predictions for similar inputs, as shown in Fig. \ref{fig:5}. More specifically, the trained model should output the same predictions for classification or  equivariant predictions for segmentation  when fed perturbed or transformed input. To this end, methods of this category learn to minimize the difference in predictions of
passing perturbed or transformed versions of a training sample through the deep network, aiming to obtain a model of better generalization ability.  The conceptual idea of consistency regularization in both single-model architecture and dual-model architecture is shown in Fig. \ref{fig:5}.  Cui \textit{et al.} \cite{cui2019semi} adapted the mean teacher model \cite{tarvainen2017mean}, an improved  teacher-student self-training strategy that also considered consistency regularization,  to semi-supervised  brain
lesion segmentation. Specifically, they minimized the differences between the predictions of
the teacher model and the student model for the same input  under different noise perturbations. Yu \textit{et al.} \cite{yu2019uncertainty} further introduced uncertainty estimation into the mean teacher learning framework.  Li \textit{et al.} \cite{li2020transformation} introduced geometric-transformation consistent loss, which was integrated into the mean teacher learning framework \cite{tarvainen2017mean} and applied to the semi-supervised segmentation of sin lesion, optic disc, and liver tumor. In the mean teacher framework, Zhou \textit{et al.} \cite{zhou2020deep} encouraged the predictions from the teacher and student networks to be consistent in both feature and semantic level
under small perturbations. They applied their model to semi-supervised instance segmentation of cervical cells.  In the teacher-student paradigm, Fotedar \textit{et al.} \cite{fotedar2020extreme} further considered  consistency under extreme transformations, including a diverse set of intensity-based, geometric, and image mixing transformations, and conducted semi-supervised lesion segmentation and retinal vessel segmentation from skin and fundus images, respectively. Liu \textit{et al.} \cite{liu2020semi} explicitly enforced the consistency of relationships among different samples under perturbations in the teacher-student framework. Instead of using the self-training strategy (i.e., the teacher-student paradigm), Bortsova \textit{et al.} \cite{bortsova2019semi} enforced transformation consistency on both the labeled and unlabeled data within a Siamese network and achieved state-of-the-art performance on chest X-ray segmentation. A similar idea has been adopted  in \cite{laradji2020weakly} to weakly supervised segmentation of covid-19 in CT images.  For semi-supervised medical image segmentation, Peng \textit{et al.} \cite{peng2020mutual} further employed mutual-information-based clustering loss to explicitly enforce prediction consistency between nearby pixels in the unlabeled images and random perturbed unlabeled images. Fang and Li \cite{fang2020dmnet} developed a convolutional network with two decoder branches of different architectures  and minimized the difference between soft masks generated by the two decoders. They applied their method to kidney tumor  and brain tumor segmentation and showed promising results.

      \begin{figure}[t]
    \centering
    \includegraphics[width=0.42\textwidth]{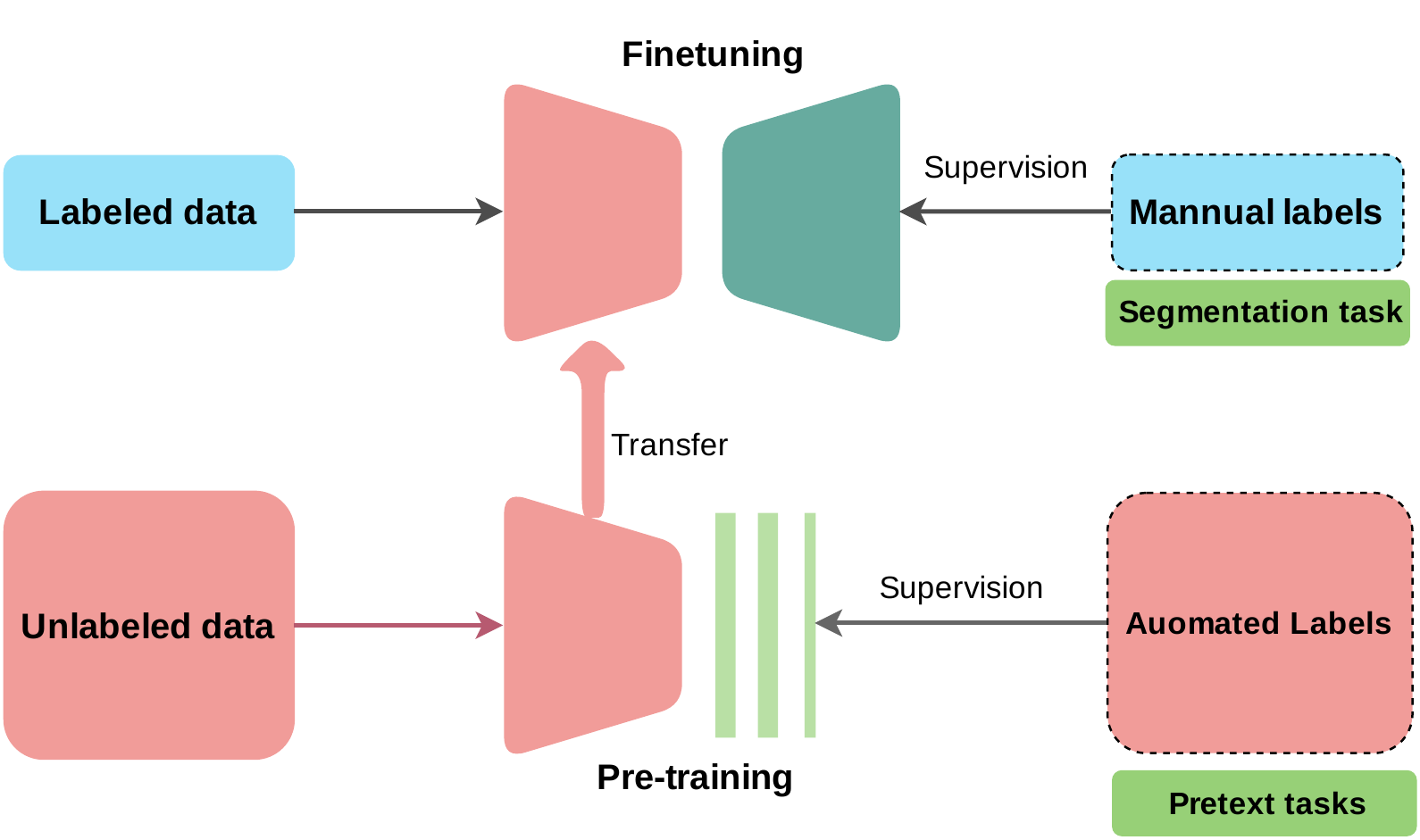}
    \caption{ The basic workflow of self-supervised semi-supervised segmentation. }
  \label{fig:6}
    \end{figure}
 \subsection{Self-supervised learning } \label{subsec: Self}
Self-supervised learning \cite{doersch2015unsupervised,noroozi2016unsupervised,larsson2016learning} (as shown in Fig. \ref{fig:6}),  a form of unsupervised learning,  has been widely used to explore unlabeled data and has shown soaring performance on representation learning through discovering data-inherent patterns.  Self-supervised learning leverages the unlabeled data with automatically generated supervisory signal and benefits the downstream tasks by self-supervised model pretraining.
%solves the given task with limited supervision by automatically generating} some kind of supervision and pretraining a model.
 Then, the pretrained model and the leaned features are adapted to the target tasks of interest. Therefore,  self-supervised learning aims to obtain a good representation of the training data without using any manual label.   A popular solution is to learn useful features by introducing various pretext tasks, such as Jigsaw puzzles \cite{noroozi2016unsupervised}, rotation prediction \cite{gidaris2018unsupervised}, inpainting \cite{pathak2016context}, colorization \cite{larsson2016learning}, relative position \cite{doersch2015unsupervised}, and a combination of a number of them, for the networks to solve, as demonstrated in Fig. \ref{fig:6}. In this way,  unlabeled training data can also be leveraged to acquire generic knowledge  under different concepts, which can be transferred to various downstream tasks.  These pretext tasks share one common property: labels for the pretext task can be automatically generated based on a certain degree of image understanding. For semi-supervised medical image segmentation, Li \textit{et al.}  \cite{li2020self} proposed generating pseudo-label by recurrently optimizing the neural network with a self-supervised task, where Jigsaw puzzles were used as the pretext task. Tajbakhsh \textit{et al.} \cite{tajbakhsh2019surrogate} used three pretext tasks, i.e.,  rotation, reconstruction, and colorization, to pre-train a deep network for different medical image segmentation tasks in the context of having limited quantities of labeled training data.   Taleb \textit{et al.} \cite{taleb20203d} extended five different pretext tasks, including  contrastive predictive coding, rotation prediction, Jigsaw puzzles, relative
patch location, and exemplar networks to 3D context, and showed  competitive results on brain
tumor segmentation from 3D MRI and pancreas tumor segmentation from 3D
CT images. In \cite{taleb2019multimodal},  Taleb \textit{et al.}  introduced a multimodal puzzle task to pretrain a model from multi-modal  images, which was then finetuned on a limited set of labeled data for  the downstream segmentation task.

  \subsection{Adversarial learning }
  Generative adversarial learning was introduced by  Goodfellow \textit{et al.} \cite{goodfellow2014generative}, which involves training two subnetworks, one  serves as a discriminator that aims to identify whether a sample is drawn from true data or generated by the
generator, and
the other as a generator that aims to generate samples that are not distinguishable by the
discriminator. The generator and discriminator are trained as a minimax two-player game. Adversarial training has been used in many applications, including fully-supervised  image segmentation \cite{luc2016semantic,yang2017automatic,han2018spine}, semi-supervised segmentation \cite{hung2019adversarial,souly2017semi,he2019dpa,han2020semi,fang2020dmnet,zheng2019semi,dai2017good,li2020shape}, and domain adaptive segmentation \cite{kamnitsas2017unsupervised,dou2018unsupervised,peng2020unsupervised,dou2019pnp,wang2019patch,chen2020unsupervised}. For semi-supervised segmentation tasks, a straightforward strategy \cite{nie2018asdnet,zhang2017deep} is to   augment the standard segmentation (the generator) with  a discriminator network designed to  distinguish between the predicted segmentation and the ground truths and choose reliable pseudo labels on the unlabeled data.
Zhang \textit{et al.} \cite{zhang2017deep} applied   adversarial learning to biomedical image segmentation with a
model  consisting of two subnetworks: a segmentation network (generator) to conduct segmentation and an evaluation network (discriminator)  to assess segmentation quality. Han \textit{et al.} \cite{han2018spine} introduced Spine-GAN, a recurrent Generative Adversarial
Network, to segment multiple spinal structures from MRIs.
 %Zhou \textit{et al.} also utilized adversarial learning and used  a discriminating network to
For semi-supervised medical image segmentation, Nie \textit{et al.} \cite{nie2018asdnet} followed a similar strategy introduced in \cite{hung2019adversarial} and utilized an adversarial network to select the trustworthy
regions of unlabeled data to train the segmentation network. Generative adversarial learning has also been used as a data-space solution to the small data problem by directly  synthesizing more realistic looking data.
 For the segmentation of unpaired multi-model cardiovascular volumes with limited training data, Zhang \textit{et al.} \cite{zhang2018translating} utilized  a cycle-consistent adversarial network for training a  cross-modality synthesis model, which can synthesize realistic looking 3D images. Cross-modality shape-consistency was enforced  to guarantee the shape
invariance of the synthetic images.

\subsection{Few-shot segmentation}
Few-shot segmentation (FSS) \cite{shaban2017one} aims at learning a model on  base semantic classes but performing
segmentation on novel semantic class with only $k$ labeled images (i.e., $k$-shot) of this unseen
class without retraining the model. The  $k$ image-label pairs for the
new class are typically referred to as the \textit{support set}. Given the \textit{support set}, FSS predicts  a binary mask of the novel class for each \textit{query image}. It is noteworthy that, in FSS, the base classes for model training  are assumed to have sufficient labeled training data, and  the novel class, i.e., the testing class, is not seen by the model during training. Although  few-shot learning  has shown promising performance for classification and  detection, its application in segmentation is immensely challenging due to the need for pixel-wise prediction.  A comprehensive review of few-shot learning (FSL) \cite{fei2006one} has been provided in \cite{wang2020generalizing}. The application of FSL to semantic segmentation of natural images was initially introduced in \cite{shaban2017one}. Rather than fine-tuning  the pre-trained model on the few support set as \cite{caelles2017one}, Shaban \textit{et al.}  \cite{shaban2017one} introduced a two-branched approach, where  the first branch takes the labeled image as input and predicts  in a single forward pass a
set of  parameters, which are used by the second branch to generate a prediction for a query image. It is noteworthy that  fine-tuning a large network on a very small support set is prone to overfitting. Roy \textit{et al}. \cite{roy2020squeeze} considered the few-shot segmentation of organs from medical volumetric images, where only a few annotated slices are available.  Following the two-branch paradigm in \cite{shaban2017one}, they introduced  strong interactions  at multiple locations between the two branches by using Channel Squeeze \& Spatial Excitation  modules \cite{roy2018concurrent,hu2018squeeze}, which is different from the one interaction at the final layer in \cite{shaban2017one}.
Ouyang \textit{et al.} \cite{ouyang2020self} introduced a superpixel-based self-supervision technique for few-shot segmentation of medical images and showed the promising ability of generalization to unseen semantic classes.

\subsection{Summary}
In previous sections, we summarize popular techniques for semi-supervised segmentation of medical images. In summary, these methods address three crucial problems: 1) how to learn a reliable  model from just a few labeled data without overfitting, 2) how to make the best use of the unlabeled data, and 3) how to use domain knowledge to learn a robust model with better generalization. Note that the three problems are not independent. The first problem can be easier to address when additional unlabeled data are available, or specific domain knowledge can be exploited.

 The first problem can be partially  addressed by data augmentation, curriculum learning, and transfer learning. Data augmentation is a simple yet effective data-space solution and artificially augment the labeled data. However, recent methods also exploit unlabeled data to capture real variations of the data. Data curriculum learning works on the data space by taking advantage of human knowledge about the training data. However, this method is not always effective.
 The transfer learning relies on the availability of external large benchmark datasets for model pretraining and can be regarded as a model-space solution. However, the effectiveness of the transfer learning, that is, adapting  the pretrained network to the current dataset, depends on the nature of the current dataset, such as the similarity of the benchmark dataset and the current dataset, and the size of the current dataset. Generally, when the similarity of the two datasets is high, and the size of the current dataset is small, the performance gain is significant. When transferring from natural image benchmarks to medical datasets,  where the data similarity is relatively low, the benefit of transfer learning is not always significant \cite{raghu2019transfusion}. Thus, pretraining  on relevant domains and applying to the current domain with supervised  or semi-supervised training, known as Domain Adaptation \cite{ben2010theory} or Domain generalization,  has  received growing attention. Please refer to \cite{toldo2020unsupervised} for comprehensive reviews of domain adaptation for semantic segmentation.

There are more methods that leverage unlabeled data, including the self-training, consistency regularization, adversarial learning, and self-supervised learning. The self-supervised learning strategy also follows the "pretrain, fine-tune" pipeline, but conducts model pretraining on the current unlabeled data in an unsupervised or self-supervised style, which is different from the supervised pretraining in transfer learning. In contrast, the consistency regularization strategy introduces unsupervised losses on the unlabeled data, and the unsupervised losses are jointly learned with the supervised loss on the labeled data. The self-training directly augments the labeled data through pseudo-labeling the unlabeled data.

There are  many  types of  domain knowledge, such as anatomical
priors by shape or atlas modeling, data or task priors by  curriculum learning and transfer learning, and so on.  Incorporating domain knowledge has proven to be effective in regularizing the model training  and especially valuable in medical image segmentation. The few-shot learning aims to generalize from a few labeled examples with prior knowledge. Thus, it can help  relieve the burden of data collecting and annotation and help learn from rare cases, which is crucial for biomedical applications.

%\begin{table}
%% table caption is above the table
%\caption{Comparison of different methods for mitochondria segmentation on EPFL dataset. \hl{The evaluation results under both class-level measures, i.e., DSC and JAC, and instance-level measures, i.e., AJI and PQ, are reported.}  }
%\centering
%\label{tab:1}      % Give a unique label
%% For LaTeX tables use
%  \setlength{\tabcolsep}{1.2mm}
%\begin{tabular}{lccccc}
%\hline\noalign{\smallskip}
%Approaches &Segmentation Task&Modality &Method\\
%\noalign{\smallskip}\hline\noalign{\smallskip}
%Baur \textit{et al.} \cite{baur2017semi} & MS Lesion  & MR&Auxiliary manifold embedding&\\
%Nie \textit{et al.} \cite{nie2018asdnet} & Postate & MR&Confidence network and region attention&\\
%Ross \textit{et al.} \cite{ross2018exploiting} &Medical instrument& Endoscopic video&Self-supervised learning&\\
%Chartsias \textit{et al.} \cite{chartsias2018factorised} &Myocardial segmentation& MR&&\\
%Yu \textit{et al.} \cite{yu2019uncertainty} & Left Atrium& MR&Teacher-student learning&\\
%Peng \textit{et al.} \cite{peng2020deep} & Cardiac/Grey matter/Spleen& MR&Co-training&\\
%Kervadec \textit{et al.} \cite{kervadec2019curriculum} & left ventricle & MR&Curriculum learning &\\
%Li  \textit{et al.} \cite{li2020self} &  & & Uncertainty estimation with Self-supervised learning&\\
%Bai  \textit{et al.} \cite{bai2017semi} & Heart & CMR &  Self-supervised learning&\\
%\noalign{\smallskip}\hline
%\end{tabular}
%\end{table}
\section{Partially-supervised Segmentation} \label{sec:4}
While semi-supervised segmentation addresses the scenario that a small subset of the training data is fully annotated, partially-supervised segmentation refers to  more challenging cases wherein partial annotations are available for all  examples  or  a subset of examples. Obviously, a model that requires only partial annotations will further reduce the workloads of manual labeling. However, this problem is more challenging than semi-supervised learning.

\textbf{Volume segmentation with sparsely annotated slices}. For 3D medical image segmentation,   uniformly sampled slices with annotations were used in \cite{cciccek20163d,bai2018recurrent,zhang2019sparse,bitarafan20203d,zheng2020cartilage} to train a 3D deep network model by assigning a zero weight to unannotated  voxels in the loss function.  Bai \textit{et al.} \cite{bai2018recurrent} performed label propagation from annotated slices to unannotated slices based on non-rigid registration and introduce an exponentially weighted loss function for model training. Bitarafan \textit{et al.} \cite{bitarafan20203d} considered a partially-supervised segmentation problem where only one 2D slice on each volume in the training data was  annotated. They addressed this problem with a self-training framework that alternatively generates pseudo labels and updates the 3D segmentation model. Specifically, they utilized the registration of consecutive 2D slices to propagate labels to unlabeled voxels. To segment 3D medical volumes with sparsely annotated 2D slices, Zheng \textit{et al.} \cite{zheng2020cartilage} utilized uncertainty-guided self-training to gradually boost the segmentation accuracy.
  Before training segmentation models with sparsely annotated slices, Zheng \textit{et al.}  \cite{zheng2020annotation} first
identified  the most influential and diverse slices for
manual annotation with a deep network.  After manual annotation of the selected slices, they conducted segmentation with a self-training strategy. Wang \textit{et al.} \cite{wang2020ct} considered a 3D image training dataset with mixed types of annotations, i.e., image volumes with a few annotated consecutive slices, a few sparsely annotated slices, or  full annotation. Under the  self-training framework, they  iteratively generated pseudo labels and updated the model with augmented labeled data. To take advantage of the incomplete annotated data, they introduced a hybrid loss, including a boundary regression loss on labeled data and a voxel classification loss on both labeled data and unlabeled data.

 \textbf{Segmentation with partially annotated regions.} As  sparse annotations, partially annotated regions or scribbles  can provide the location and label information at a few pixels or partial regions. They have been  widely used in segmentation tasks \cite{boykov2001interactive,lin2016scribblesup,tang2018regularized,tang2018normalized,cheng2020self}, especially in the context of interactive segmentation \cite{rother2004grabcut,majumder2019content,han2020deep,wang2018deepigeos,wang2020uncertainty,zhang2020weakly}, where users give feedback to refine the segmentation iteratively. Scribbles  have  been recognized as a user-friendly way of interacting for both natural and medical image segmentations \cite{rother2004grabcut,lin2016scribblesup,boykov2001interactive,peng2014liver}. While  scribble-supervised segmentation was usually addressed  by optimizing a
graphical model \cite{boykov2001interactive} or a variational model \cite{chan2006algorithms}, tackling this problem with deep networks has also been a hot topic.

  Since medical images usually suffer from  low tissue contrast, fuzzy boundaries, and image artifacts such as noise and intensity bias, an interactive strategy is also valuable for medical image segmentation. Zhang \textit{et al.} \cite{zhang2020interactive} considered interactive medical image segmentation via a point-based interaction, where the physician should click on rough central points of the objects for each testing image. For MR image segmentation,  Wang \textit{et al.} \cite{wang2018interactive} employed scribbles as user interaction to fine-tune coarse segmentation in the context of deep learning. Zhou \textit{et al.} \cite{zhou2019interactive} introduced an interactive editing network trained using simulated user interactions to refine  the existing segmentation. Liao \textit{et al.} \cite{liao2020iteratively} proposed to solve the  dynamic
process of iteratively interactive image segmentation  of 3D medical images with multi-agent reinforcement learning, where they treated each voxel as an agent with  shared behaviors.

 To reduce the annotation effort in the context of interactive segmentation, especially for instance segmentation, researchers have explored methods  to suggest annotations \cite{yang2017suggestive,wang2018deep,belharbi2020deep,sourati2019intelligent} or select informative samples \cite{mahapatra2018efficient}. A promising solution is  \textit{active learning},  the process of selecting the examples or regions that need to get human labels.  In this way, we can obtain a model that achieves the desired accuracy faster and  with low annotation cost. The flowchart of  active learning for interactive segmentation
on a conceptual level is shown in Fig. \ref{fig:3} (b).  A comparison with self-training, which explores no human interaction,  is also demonstrated. Yang \textit{et al.} \cite{yang2017suggestive} combined the deep network model with active learning to identify the most representative and uncertain areas for annotation.  In instance-level segmentation of pulmonary nodules, Wang \textit{et al.} \cite{wang2018deep} utilized active learning to  overcome the annotation bottleneck by querying the most confusing unannotated instances for manual annotation. The most confusing unannotated instances were identified with high-uncertainty.  For breast cancer segmentation on immunohistochemistry images,  Sourati \textit{et al.} \cite{sourati2019intelligent} introduced a new active learning method with Fisher information for deep networks to identify a small number of the most informative  samples to be manually annotated. To achieve a rapid increase in the segmentation performance, Shen \textit{et al.} \cite{shen2020deep} designed three  criteria, i.e.,  dissatisfaction, representativeness, and diverseness, in the framework of active learning  to select an informative subset for labeling, which can obviously reduce the cost of annotation.  Given an initial segmentation, Wang \textit{et al.} \cite{wang2020uncertainty} used uncertainty estimation to identify a subset of slices that require user interactions.

 Rather than performing interactive segmentation, Lin \textit{et al.} \cite{lin2016scribblesup}  directly used the given sparse scribbles as the supervision to train a deep convolutional network for natural image segmentation.
 Tang \textit{et al.} \cite{tang2018regularized} investigated partially-supervised segmentation with scribbles as annotations and introduced several regularized losses, including CRF \cite{boykov2001interactive} loss, high-order normalized cut loss, and kernel cut loss in the context of deep convolutional networks. In \cite{tang2018normalized},  Tang \textit{et al.} further introduced normalized cut loss. Ji \textit{et al. } \cite{ji2019scribble} investigated the segmentation of brain tumor substructures with whole tumor/normal brain scribbles and the image-level labels as supervision.  To capture fine tumor boundaries, they augmented the segmentation network with a dense CRF \cite{krahenbuhl2011efficient} loss. For 3D instance segmentation in medical images, Zhao \textit{et al.} \cite{zhao2018deep} considered a mix of 3D bounding boxes for all instances and  voxel-wise annotation for a small fraction of the instances. They addressed this problem with a cascade of two stages: an instance detection stage with bounding-box annotations and an instance segmentation stage with full  annotations for a small number of instances.

   For semantic segmentation of emphysema with
both annotated and unannotated areas in training data, in the self-training framework, Peng \textit{et al.} \cite{peng2019semi} utilized the similarities of deeply learned features between labeled and unlabeled areas to guide the label propagation to  unannotated areas. Then, the
 selected  regions with confident pseudo-labels are used to  enrich the training data.  For the segmentation of  cancerous regions in gigapixel whole slide
images (WSIs), Cheng \textit{et al.} \cite{cheng2020self} considered a partially labeled scenario, where only partial cancer regions in
WSIs were annotated by pathologists due to time constraints or misinterpretation. To tackle this problem, they integrated the idea of   the teacher-student learning paradigm and self-similarity learning to enforce nearby
patches in a WSI to be similar in feature space.  A similar prediction ensemble strategy was used to generate pseudo labels, which were used to filter out noisy labels.

 Dong \textit{et al.} \cite{dong2020towards} considered neuron segmentation from macaque brain images, where both central points and rough masks were used as the supervision.  Zheng \textit{et al.} \cite{zheng2020weakly} proposed to use boundary scribble as the weak supervision for tumor segmentation, where boundary scribbles are  coarse lesion edges. While boundary scribbles provide  both location information about the  lesions and   more accurate boundary information than bounding boxes, they still lack  information about accurate boundaries. For cell segmentation with scribble annotations,  Lee and Jeong \cite{lee2020scribble2label} proposed to generate reliable labels through the integration of  pseudo-labeling and label filtering in the mean teacher framework \cite{tarvainen2017mean}.
 \begin{figure}[!t]
    \centering
   % \subfloat[EM image]{\includegraphics[width=0.155\textwidth]{test015.pdf}} \hspace{0.01mm}
%    \subfloat[Semantic seg.]{\includegraphics[width=0.155\textwidth]{GT3.pdf}} \hspace{0.01mm}
 \subfloat[EM image]{\includegraphics[width=0.22\textwidth]{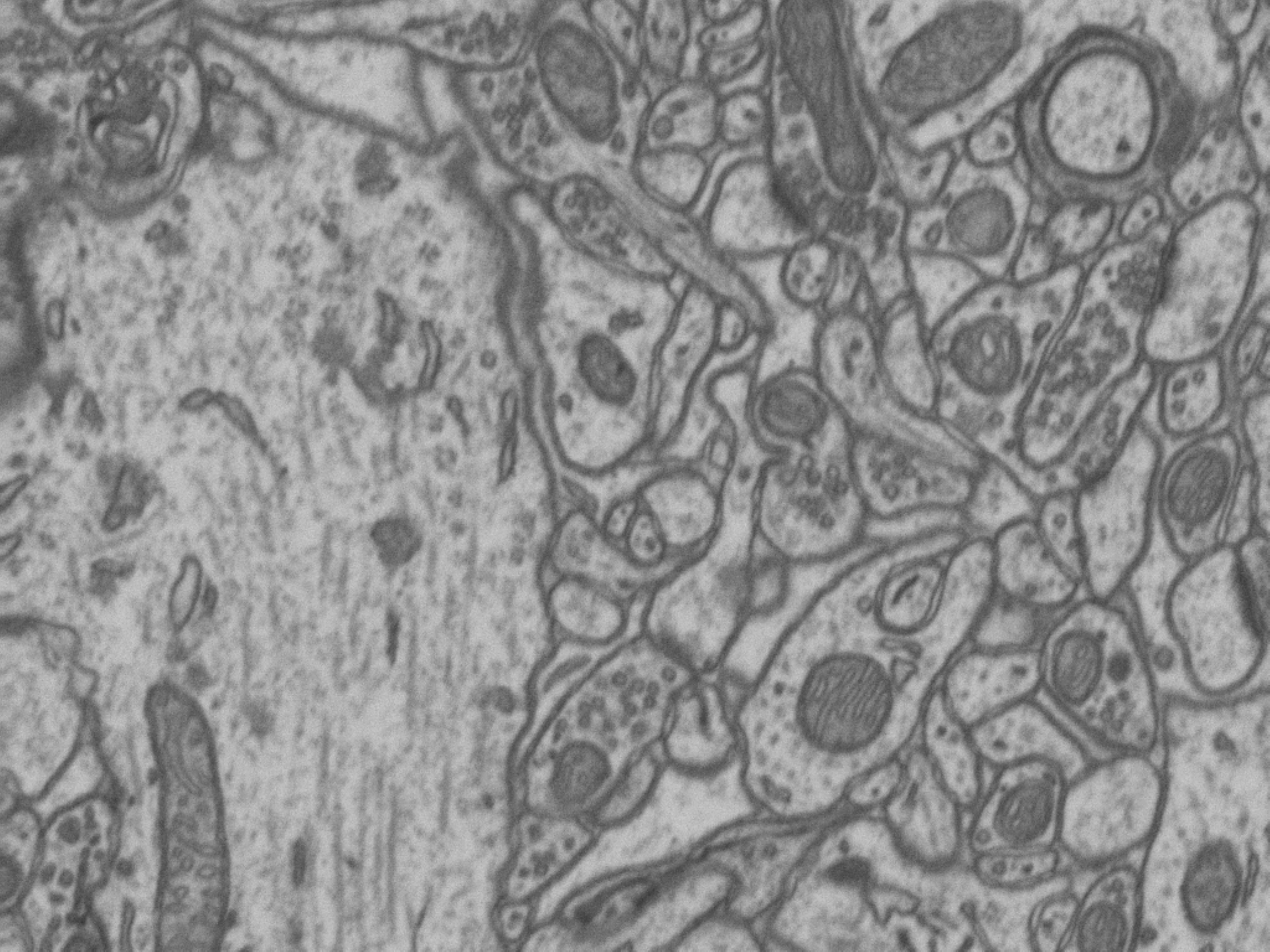}}\hspace{0.001mm}
      \subfloat[Pixel-wise annotation]{\includegraphics[width=0.22\textwidth]{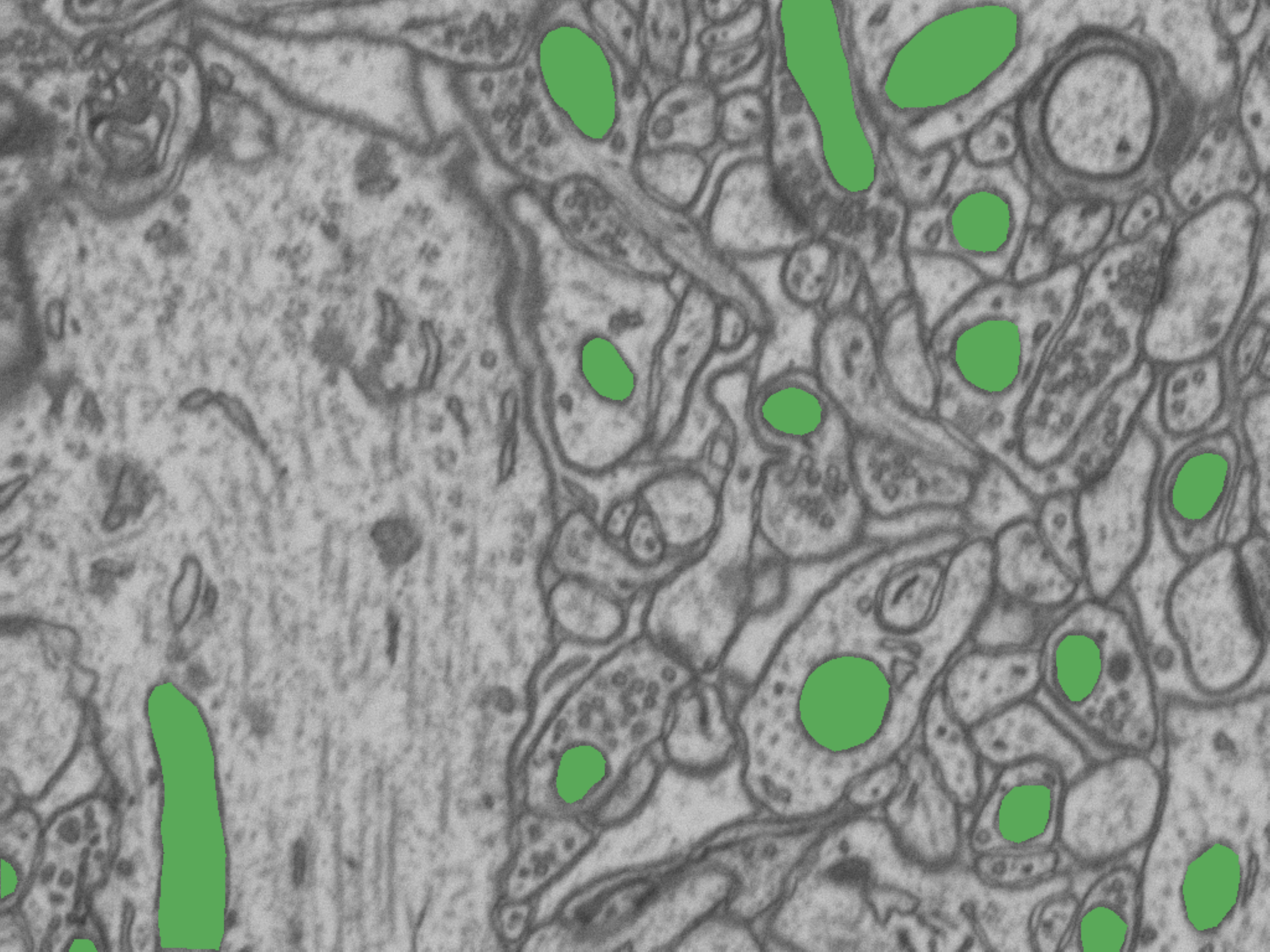}}\\
     % \subfloat[Instance seg.]{\includegraphics[width=0.155\textwidth]{instance3.pdf}
    \subfloat[Point annotation]{\includegraphics[width=0.22\textwidth]{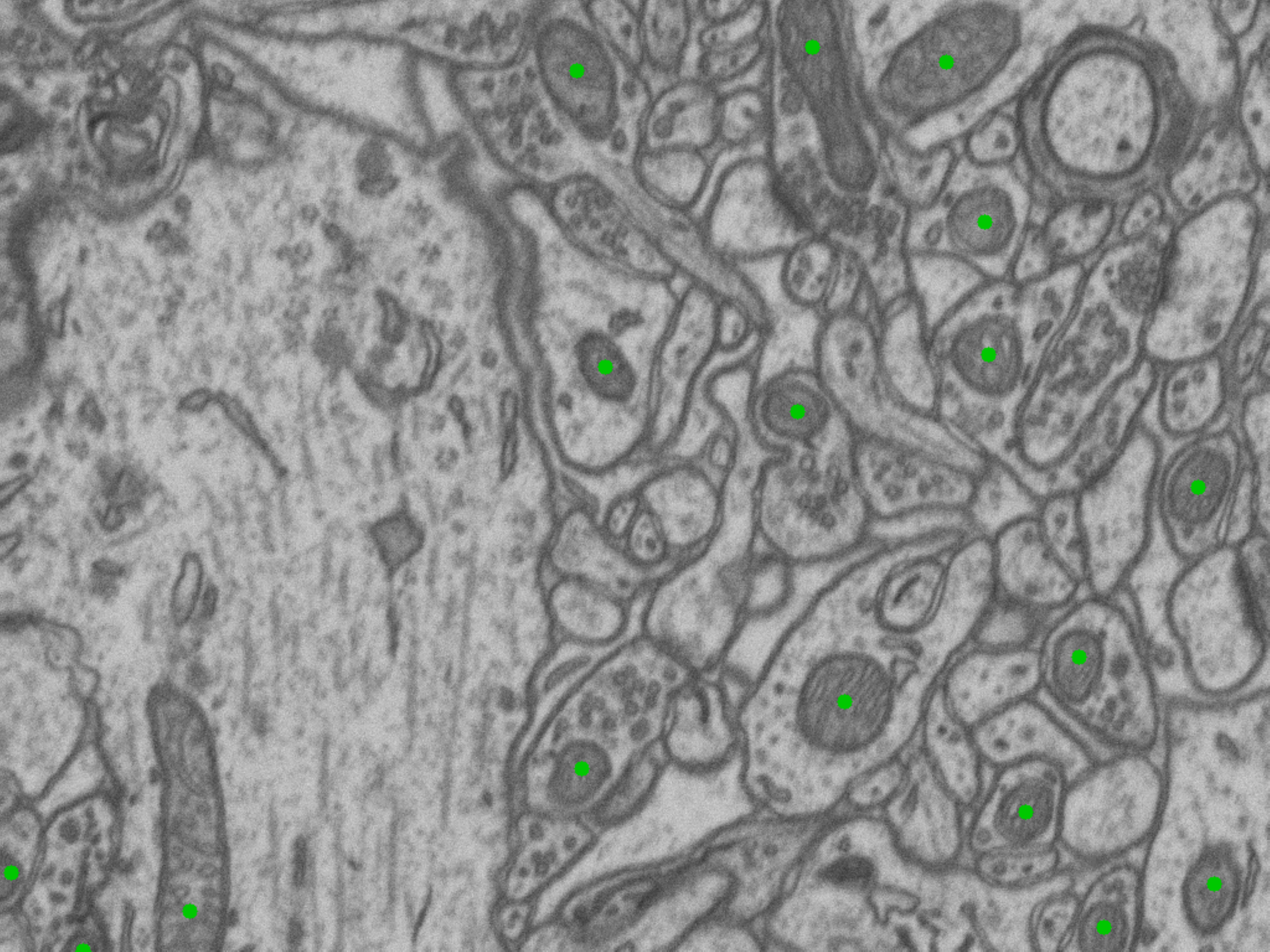}} \hspace{0.001mm}
    \subfloat[Bounding-box annotation ]{\includegraphics[width=0.22\textwidth]{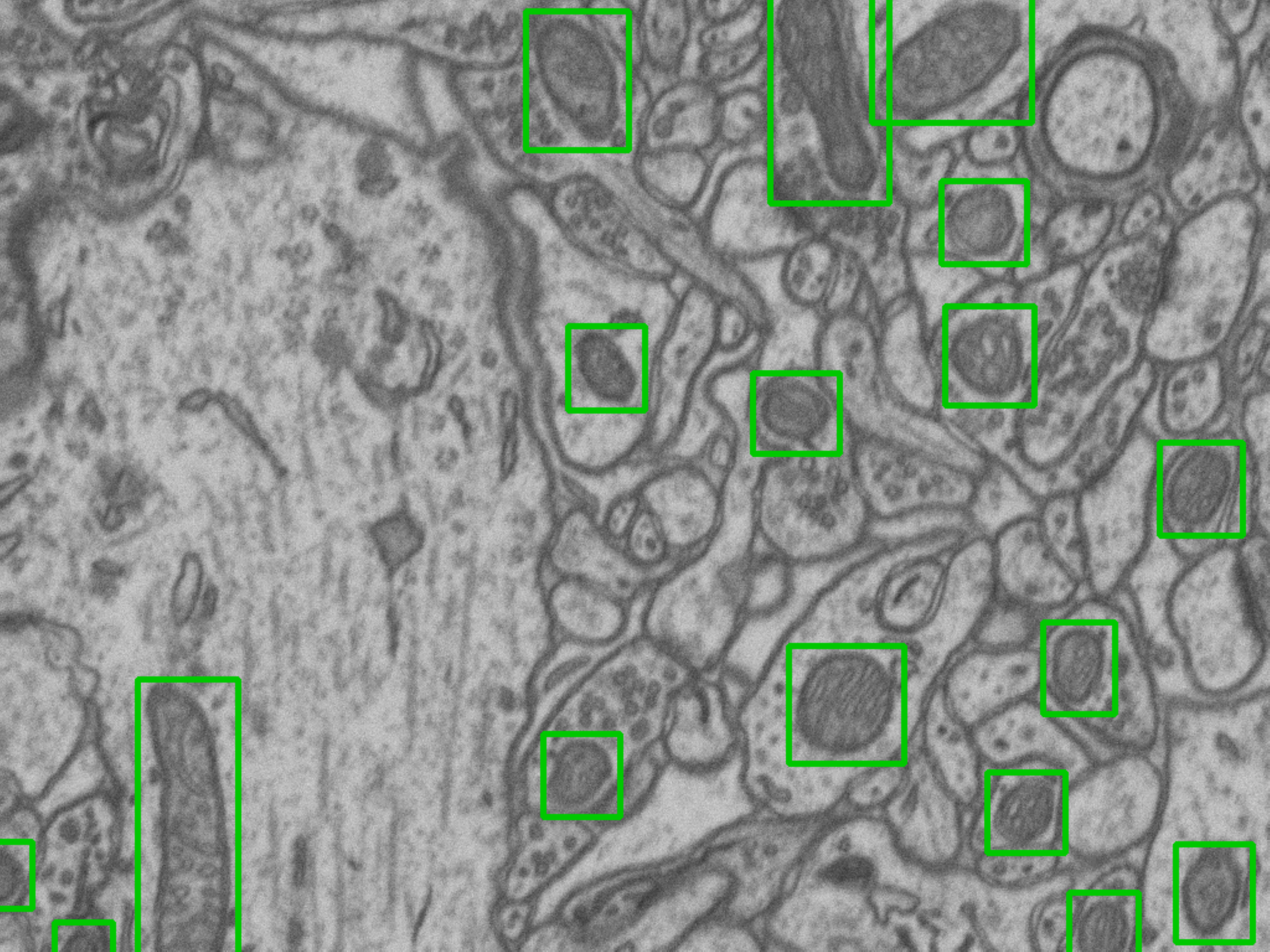}}
 %   \subfloat[a]{\includegraphics[width=0.48\textwidth]{StudentTeacher10.png}}
    \caption{An illustration of pixel-wise annotation, point annotation and bounding-box annotation for mitochondria segmentation from  electron microscopy (EM) images. }
  \label{fig:00}
  \end{figure}
\textbf{Segmentation with point annotations.} An extremely sparse annotation is point annotation \cite{bearman2016s}, which is labeling only one point in each object as examplified in Fig. \ref{fig:00}.
Point annotations \cite{bearman2016s,can2018learning,lejeune2018iterative,qu2020weakly,qu2020nuclei} have also been considered to reduce the cost of manual labeling, which is especially useful for multiclass segmentation and instance-level segmentation.  Point annotation
is one of the fastest ways to label objects. As has shown in  \cite{bearman2016s}, point annotation is significantly cheaper than dense pixel-level annotation. Despite its cost-efficiency, point annotations are  extremely sparse annotations and  only contain location information. Thus, most  studies have utilized point annotations on object detection and counting  tasks \cite{laradji2018blobs,liu2019point,gao2020renal}, such as cell detection \cite{kainz2015you,gao2020renal} and nuclei detection \cite{zhou2018sfcn}.  Yoo \textit{et al.} \cite{yoo2019pseudoedgenet} investigated nuclei segmentation
 with point annotations. Since point annotations do not contain nucleus boundary information, they augmented  the segmentation network with an auxiliary network for edge detection with  the Sobel filtered prediction map of the segmentation network as the supervision signal.  To segment mitosis from breast histopathology images with centroid labels, Li \textit{et al.} \cite{li2019weakly} expanded the single-pixel label to a novel label with concentric circles, where  the inside circle was regarded as a mitotic region, and the regions outside the outer ring were regarded as  non-mitosis. They introduced a concentric loss to make the segmentation network be trained only with the estimated labels in the inside circle and outside the outer circle.
 For nuclei segmentation, Qu \textit{et al.} \cite{qu2019weakly,qu2020weakly} addressed a more challenging case, where only sparse points annotation, i.e., only a
small portion of nuclei locations in each image, were annotated with center points. Their method consists of two stages, the first stage conducts nuclei detection with a self-training strategy, and the second stage performs semi-supervised segmentation with pseudo-labels generated with Voronoi diagram and k-means clustering. The dense CRF loss was utilized in training to  refine the segmentation.
%For nuclei instance segmentation, Yen \textit{et al.} proposed to use pseudo-label strategy  with
%point annotations.

\textbf{Multiclass segmentation from multiple few-class-labeled datasets}.
Segmentation of various anatomical  structures from medical images, such as multi-organ segmentation, is a fundamental problem in medical image analysis and downstream clinical usage. However, besides the cost of data collection, obtaining  sufficient multiclass annotations on a large  dataset in itself is a labor-intensive and often impossible task. In contrast, there are various  annotated datasets from different
medical centers for their own clinical and research purposes but with missing annotations for several classes, or even  only single-class annotations  available. Most  public available medical image data are designed and annotated  for  specific clinical and research purpose, such as Sliver07 \cite{heimann2009comparison} for evaluation of liver segmentation \cite{peng20153d,hu2016automatic,peng2014liver,lu2017automatic}, LiTS \cite{bilic2019liver} for evaluation of liver-tumor segmentation,  KiTS \cite{heller2019kits19} for evaluation of kidney-tumor  segmentation \cite{heller2019state}, LUNA \cite{setio2017validation} for lung nodule analysis \cite{xie2019semi,cao2020two}, and BRATS \cite{menze2014multimodal} for brain tumor analysis \cite{wang2017automatic,luo2020hdc}. Thus, a significant challenge is how to learn a universal multi-class  segmentation model from multiple partially annotated datasets with missing annotated classes.

Zhou \textit{et al.} \cite{zhou2019prior} first considered a partially-supervised
multi-organ segmentation problem where a small fully labeled dataset and several partially-labeled datasets are available. They  developed a prior-aware neural network that explicitly incorporated anatomical priors on abdominal organ sizes as domain-specific knowledge to guide the training process. Dmitriev \textit{et al.} \cite{dmitriev2019learning} further removed the need for the fully labeled data  and investigated the problem of multi-class (e.g.,multi-organ) segmentation from  single-class (e.g., single-organ) labeled datasets \cite{zhou2019prior}. They proposed to condition a single convolutional network  for  multi-class segmentation with non-overlapping single-class datasets for training. Concretely, they  inserted the conditional information as an intermediate activation between
convolutional operation and the activation function. Huang \textit{et al.} \cite{huang2020multi} tackled the problem of partially-supervised multi-organ segmentation in the co-training framework, where they  collaboratively trained multiple networks, and each network was taught by other networks on un-annotated organs. Yan \textit{et al.} \cite{yan2020learning} proposed to develop a
universal lesion detection algorithm to detect a comprehensive variety of lesions from multiple datasets with partial labels. Specifically, they introduced intra-patient lesion matching and cross-dataset lesion mining to address missing annotations and  utilized feature sharing, proposal fusion, and annotation mining to  integrate different datasets.    Shi \textit{et al.} \cite{shi2020marginal} addressed partially-supervised multi-organ segmentation with two novel losses: the marginal loss that aims to merge all unlabeled organ pixels with the background label, and  the exclusion loss that constrains multi-organs exclusive.  To segment multi-organ and tumors from multiple partially labeled datasets,  Zhang \textit{et al.} \cite{zhang2020dodnet} proposed an encoder-decoder network with a single but dynamic head, in which the kernels are generated adaptively by a controller, conditioned on both the input image and assigned task. Dong \textit{et al.} \cite{dong2020towards} considered a special case  where full labels for all classes are not available on the whole training set, but labels of different  classes are available on different subsets. They addressed this problem with a data augmentation strategy by exploiting
the assumption that patients share anatomical similarities.

\textbf{Summary.} Reducing annotation cost essentially echoes the real-world environments, where the annotations are often incomplete or even sparse. This section covers four types of partial annotations, including partially annotated slices for 3D images,  partially annotated regions, point annotations, and multiple few-class-labeled datasets. As shown in previous sections, most methods  to address these scenarios are based on self-training and regularization techniques. When collecting annotated data with a human in the loop, suggestive annotations can significantly reduce the annotation effort, mostly when extensive scale data should be annotated, or a large number of instances need annotations.
Thus, a critical question is  which data samples or image regions should be  selected for annotations to achieve high-quality performance faster. This active learning paradigm, as exemplified  in Fig. \ref{fig:3} (b), has been an active research field, where more efforts are still needed.

\section{Inaccurately-supervised segmentation} \label{sec:5}
Segmentation with inaccurate or imprecise annotations refers to the scenario where the ground truth labels are corrupted with (random, class-conditional or  instance-conditional \cite{menon2018learning,cheng2020learning}) noises, thus also referring to \textit{noisy label learning} \cite{natarajan2013learning,angluin1988learning}. Imprecise boundaries, and mislabeling are also inaccurate annotations.
%It is noteworthy that partial labels such as annotated slices, scribbles, and points, can be regarded as missing annotation, where the labels of positive classes have varying degrees of corruption.
Moreover,  bounding-box annotations can be treated as annotations with inaccurate boundaries and mislabeled regions. Note that, as shown in \cite{heller2018imperfect},  boundary-localized errors are more challenging than random label errors.

\textbf{Learning from noisy labels} has recently drawn much attention in many applications, including medical image analysis \cite{karimi2020deep,le2019pancreatic,gao2017deep}. It is expensive and sometimes infeasible to obtain accurate labels, especially on medical imaging data where labeling requires
domain expertise, and annotating huge-size imaging data is inherently a daunting task. In contrast, noisy labels such as those generated by non-experts \cite{zhang2020robust} or computers \cite{zhang2020weakly} are easy to obtain. Moreover, it is impractical to manually correct the  label errors, which is not only  time-consuming bust also  requires a stronger committee of experts.
Karimi \textit{et al.} \cite{karimi2020deep} provided a review of the state-of-the-art deep learning methods (published in 2019 or earlier) in handling label noise. However, most approaches for dealing noisy (low-quality) annotations are developed for  classification \cite{dgani2018training,xue2019robust,han2018masking,algan2020deep,cao2020breast} and detection \cite{karimi2020learning}. Herein, we focus on medical image segmentation  with noise labels.

While a class of methods struggles to model and learn label noises \cite{ren2018learning,du2015modelling}, other methods choose to select confidence examples,  reducing the side effects without explicitly modeling the label noise \cite{han2018co,xue2019robust,zhu2019pick}, such as the co-teaching paradigm \cite{han2018co} and reweighting strategy \cite{ren2018learning}. To eliminate the disturbance of segmentation from inaccurate labels, Zhu \textit{et al.}  \cite{zhu2019pick}  developed a label quality evaluation strategy with a deep neural network to automatically assess the label quality. They trained the segmentation model on  examples with clean annotations.
For chest X-ray segmentation with imperfect labels, Xue \textit{et al.} \cite{xue2020cascaded} adopted a cascade strategy  consisting of two stages: a  sample selection stage, which selects the  clean annotated examples as the co-teaching paradigm, and  a label correction and model learning stage, which learns the segmentation model from both the corrected labels and original labels. To segment skin lesions from noisy annotations,  Mirikharaji \textit{et al.} \cite{mirikharaji2019learning} adopted a spatially adaptive reweighting approach to emphasize the learning from clean labels and reduce the side effect  of noisy pixel-level annotations. A meta-learning was adopted to assign higher importance to pixels.
 Shu \textit{et al.} \cite{shu2019lvc} proposed to enhance supervision of noisy labels by capturing local visual saliency features, which are less affected by supervised signals from inaccurate labels.  For noisy-labeled medical image segmentation, Zhang \textit{et al.} \cite{zhang2020characterizing} integrated  confidence learning \cite{northcutt2019confident}, which can identify the label
errors through estimating the joint distribution between the noisy
 annotations and the true (latent) annotations, into the teacher-student framework  to identify the corrupted labels at pixel-level. Soft label correction based on spatial label smoothing regularization was also adopted to generate high-quality labels.
Rather than using  fully manual annotations for vessel segmentation, Zhang \textit{et al.} \cite{zhang2020weakly}  proposed to learn the segmentation from noisy pseudo labels obtained from automatic vessel enhancement, which usually has system bias. To tackle this problem, they adopted  improved self-paced learning with online guidance from additional sparse manual annotations. The self-paced learning strategy enabled the model training to focus on easy pixels, which have  a higher chance to have  correct labels.  To minimize manual annotations, they introduced a model-vesselness uncertainty estimation for suggestive annotation. To weaken the influence  of the noise pseudo labels in semi-supervised segmentation, Min \textit{et al.} \cite{min2019two} introduced a two-stream mutual attention network with hierarchical distillation, where the multiple attention layers were used to  discover incorrect labels  and indicate potentially incorrect gradients.

\textbf{Segmentation with bounding-box annotations.}
A appealing  weak supervision is bounding-box annotations \cite{dai2015boxsup,papandreou2015weakly,khoreva2017simple}, which are easy to obtain and can yield confirmed information about backgrounds and rich information about the foreground. Moreover, the bounding box, as shown in Fig. \ref{fig:00}, can be simply represented by two  corners, which allows  light storage. Given the uncertainty of figure-ground separation within each bounding box  \cite{hsu2019weakly}, one of the core tasks for bounding-box supervised segmentation is to generate accurate pseudo-labels.
A popular pseudo-label generation approach is  Grabcut \cite{rother2004grabcut},  which iteratively estimates  the foreground and background's distributions and conducts segmentation with CRF models  such as graph cut \cite{boykov2001interactive}. The iterative strategy that alternatively updates segmentation model parameters and pseudo labels have been widely used to address bounding box annotations \cite{dai2015boxsup,rajchl2016deepcut}. In the context of natural image segmentation with deep networks,  the BoxUp model \cite{dai2015boxsup} iterated between automatically generating region proposals and training convolutional networks. For fetal brain segmentation from MR images with bounding-box annotations,  Rajchl \textit{et al.} \cite{rajchl2016deepcut} introduced DeepCut model, an extension of the  GrabCut method to estimate distributions by training a deep network classifier. Specifically, they  iteratively optimized a densely-connected CRF model and a deep convolutional network.
 Kervadec \textit{et al.} \cite{kervadec2020bounding} leveraged the classical bounding-box tightness prior \cite{lempitsky2009image} to regularize the output of deep segmentation network. Concretely,  the bounding-box tightness prior was reformulated as  a set of foreground constraints and a global background emptiness constraint, which enforced the regions outside bounding box to contain no foreground. They solved the introduced energy function with inequality constraints a sequence of unconstrained losses based on an extension of the log-barrier method.  Wang \textit{et al.} \cite{wang2020iterative} investigated the segmentation of male pelvic organs in CT from 3D bounding box annotations, which was addressed with  iterative learning of deep network model and pseudo labels. A label denoising
module, which evaluated the consistency of predictions  across consecutive iterations, was designed to  identify the voxels with unreliable labels. Zheng \textit{et al.} \cite{zheng2020weakly} proposed to use boundary scribble as the weak supervision for tumor segmentation, where boundary scribbles are  coarse lesion edges. While boundary scribbles include locations of lesions and provide more accurate boundary information than bounding boxs, they still lack accurate information about boundaries.

\textbf{Summary.} Lower the requirement of precise annotations can also significantly reduce annotation efforts. In this section, we have reviewed two types of inaccurate annotations, i.e., noisy labels and boxing-box label. While the partial annotations reviewed in Section \ref{sec:4} are reliable annotations for the positive classes (except for the background class), inaccurate annotations in this section refer to unreliable labels. For example, the noisy labels can be regarded as labels corrupted from ground truth labels; the bounding-box annotations, as shown in Fig. \ref{fig:00} (d),  contain both foreground pixels and background pixels. It is known that deep network models are susceptible to  the presentation of label  corruptions \cite{chakraborty2018adversarial,karimi2020deep}. Thus, addressing label noises has gained increasing attention in recent years and has been a popular topic in top conference venues.

%Multi Source Inexact and Inaccurate Supervision
% high level or less precise information

%\section{Weakly-supervised Segmentation}
%
%
%\cite{xu2019camel}
%global label statistics
%
%Incomplete supervision: only a subset of training data is given with labels, active learning with human intervention, semi-supervised learning without human intervention
%
%Inexact supervision: training data is only given with coarse-grained labels,  multiple instance learning
%
%Inaccurate supervision: the given labels are not always ground truth  learning with noise label
\section{Discussion and future directions}\label{sec:6}
In this section, we  discuss some ongoing or future directions of medical image segmentation with limited supervision.

\textbf{Task-agnostic versus task-specific use of unlabeled data}  Semi-supervised segmentation methods partially differ in how to leverage unlabeled data.
%In the context of semi-supervised learning,
%a core task is to exploit the unlabeled data, which includes vital information about the underlying data distribution, and typically assume to be valuable for constructing a  model with better robustness and better generalization ability.
 There are two typical ways to make use of unlabeled data: 1) the task-agnostic approach, which leverages unlabeled data through unsupervised or self-supervised pretraining, such as the self-supervised learning strategy in Sec. \ref{subsec: Self}; 2) the task-specific approach, which jointly leverages the labeled  and unlabeled data by enforcing a form of regularization, such as the consistency regularization strategy in Sec. \ref{subsec: Consistency} and self-training in Sec. \ref{subsec: selftraining}. While the task-agnostic approaches utilize the unlabeled data for  unsupervised representation learning followed by  supervised fine-tuning, the task-specific approaches use the unlabeled data to directly augment the labeled data through pseudo-labeling, or regularize the supervised model learning through consistency regularization.
Both paradigms   have shown promising results and received substantial attention in the fields of medical imaging and computer vision. Recently, an  encouraging progress in self-supervised learning  is the contrastive learning \cite{chen2020simple,he2020momentum},  which formulates the task of discriminating similar and dissimilar things in the learning model. The Momentum Contrast  model \cite{he2020momentum} with contrastive  unsupervised pretraining  outperformed its supervised pretrained counterpart in several natural image segmentation tasks. The contrastive learning strategy also has been used in medical image segmentation with limited annotations and has shown promising results \cite{chaitanya2020contrastive}. However, the gap between the objectives of the self-supervised pretraining and downstream segmentation task is non-negligible. More work in this direction is expected to push the boundaries on medical image segmentation tasks. Another promising direction is integrating the task-agnostic and task-specific approaches in an elegant way. A possible solution is introduced in \cite{chen2020big}. They  first fine-tuned the unsupervised pretrained model and then  distilled the model  into a smaller one with the unlabeled data.

% Recently, The recently introduced  , also belongs to the task-agnostic approach.
\textbf{More constructive theoretical analysis  is needed.} Although diverse strategies, such as self-supervised learning, and curriculum learning,   have been introduced and  achieved promising results, more studies are needed to identify their mechanisms. For example, it is still unclear  how to automatically design an adaptive curriculum for the given segmentation task instead of using  a predefined curriculum \cite{bengio2009curriculum} and when will curriculum-like strategies, especially data curriculum,  benefit the deep model training. Possible promising directions for automatic curriculum design may be the self-pace paradigm \cite{jiang2015self} and teacher-student paradigm \cite{matiisen2019teacher}.  Whereas  curriculum  learning has achieved success on classification and detection tasks \cite{tang2018attention,zhao2020egdcl,jimenez2019medical,oksuz2019automatic}, it has relatively limited applications on semi-supervised medical image segmentation tasks  \cite{jesson2017cased,wang2018deep}. In addition to more extensive experimental results on diverse segmentation tasks, there is also a need for theoretical guarantees on their effectiveness\cite{hacohen2019power,weinshall2020theory}, which is the foundation for its application on specific tasks. A solid foundation is also needed for segmentation methods based on self-supervised learning, especially those using consistency loss or contrastive loss \cite{arora2019theoretical}.

\textbf{Lightweight and efficient segmentation  models are favorable.}
Deep and wide models
are slow to train and, more importantly, they may easily overfit on datasets with limited annotation. Lightweight models with few parameters and few computational resource requirements are favorable for model training and deploying on computationally limited platforms, which may significantly improve the clinical application's efficiency \cite{sandler2018mobilenetv2,luo2020hdc,karakanis2020lightweight}. Two strategies are usually employed:  model compression \cite{cheng2018model} and efficient model architecture designs \cite{zhang2018shufflenet,sandler2018mobilenetv2,luo2020hdc}. Moreover, model cascade, model ensemble, maintaining multiple networks, and combining them, such as the self-training strategy,  are usually used, which inevitably increase system complexity and degrades training  efficiency \cite{luo2020hdc}.  Thus, maintaining a more simple model system is challenging future direction.

\textbf{Hyper-parameter searching is
challenging. } There are usually more hyper-parameters, such as the trade-off parameters, in segmentation methods with limited supervision. However, there are not enough labeled data for reliable hyper-parameter searching, resulting in high-variance in performance. A possible solution is using meta-learning \cite{ravi2016optimization}, the goal of which is  `learning to learn better'. In other words, meta-learning seeks to improve the learning algorithm itself with either task-agnostic or task-specific prior knowledge and thus can improve both data and computational efficiency. Thus, there is a rapid growth in
interest in meta-learning and its various applications, including medical image segmentation with limited supervision \cite{khandelwal2020domain}.  Utilizing meta-loss on a small set of labeled data has shown promising results in few-shot learning \cite{ravi2016optimization,tian2020differentiable,ren2018meta}.

\textbf{Complex label noises   are
challenging.} Label noises in real-world applications are usually a mix of several types of noises, such as class-dependent noise, instance-dependent noise, and adversarial noise, which tends to confuse models on ambiguous regions or instances. Thus, training models with the ability to tackle complex noises is valuable for real-world clinical applications.  Karimi \textit{et al. }\cite{karimi2020deep} reviewed
deep learning methods dealing with label noises for medical
image analysis, where most of the representative studies are about medical image classification. However, the challenge of dealing with label noises is particularly significant in segmentation tasks since pixel-wise labeling of large datasets is resource-intensive and requires experts' domain knowledge. The limited imaging resolution also makes the annotators difficult to identify small objects and  fuzzy boundaries. More label noises exist in  the large number of annotations produced by non-experts or automatics labeling software with little human refitments. To  analyze and address various kinds of label errors, an important thing is to construct large scale datasets with real noises, which in itself is a challenging task. Currently, most studies still use public datasets with simulated label perturbations \cite{heller2018imperfect,xue2020cascaded,zhang2020characterizing} or private datasets  \cite{shu2019lvc,karimi2020deep}. Building up public benchmarks with real noises is crucial to make further breakthroughs, especially for clinical usage.

\textbf{Learning to represent and integrate  domain knowledge is still challenging.} Although domain knowledge has  dramatically boosted  medical image segmentation methods, especially in settings with limited supervision,  the selection and representation of  prior knowledge are still challenging since it is usually highly dependent on the specific task.  Xie  \textit{et al.} \cite{xie2020survey} summarized  recent progress on integrating  domain knowledge into deep learning models for medical image analysis. Moreover, translating the original representation of prior knowledge in clinical settings to the representations  that are ready for the integration with deep networks is challenging.
\section{Conclusions}\label{sec:7}
In this review, we covered effective solutions for  the segmentation of biomedical images with limited supervision, namely, semi-supervised segmentation, partially-supervised segmentation, and inaccurately-supervised segmentation. We reviewed a diverse set of methods for these problems. For semi-supervised segmentation, we provide a taxonomy of existing methods that are with the ability to  leverage labeled data, unlabeled data, and also prior knowledge. For the task of partially-supervised segmentation, we considered segmentation with partially annotated regions, point annotations, or partially annotated slices, interactive segmentation, and multi-class segmentation from multiple partial-class-labeled datasets and shown the current technical status regarding recent solutions. For the task of inaccurately-supervised segmentation, we summarized the methods addressing noisy labels and bounding-box annotations.
%Drawing on the pros and cons of the reviewed methods as well as their connections,
We also have discussed possible future directions for further studies.

\end{document}